\documentclass[runningheads]{llncs}

\usepackage{eccv}

\usepackage{eccvabbrv}

\usepackage{graphicx}
\usepackage{booktabs}
\usepackage{xcolor} 
\usepackage{adjustbox}
\usepackage{wrapfig}
\usepackage{amsmath}
\usepackage[accsupp]{axessibility}  

\usepackage{hyperref}

\usepackage{orcidlink}

\begin{document}

\title{Text2Place: Affordance-aware Text Guided Human Placement}   

\titlerunning{Text2Place}

\author{Rishubh Parihar \and
Harsh Gupta \and
Sachidanand VS \and
R. Venkatesh Babu} 

\authorrunning{R. Parihar \textit{et al.}}

\institute{Vision and AI Lab, Indian Institute of Science, Bangalore\\
\href{https://rishubhpar.github.io/Text2Place/}{\textcolor{blue}{Project Page}}}

\maketitle
\begin{abstract}
For a given scene, humans can easily reason for the locations and pose to place objects. Designing a computational model to reason about these affordances poses a significant challenge, mirroring the intuitive reasoning abilities of humans. This work tackles the problem of realistic human insertion in a given background scene termed as \textbf{Semantic Human Placement}. This task is extremely challenging given the diverse backgrounds, scale, and pose of the generated person and, finally, the identity preservation of the person. We divide the problem into the following two stages  \textbf{i)} learning \textit{semantic masks} using text guidance for localizing regions in the image to place humans and \textbf{ii)} subject-conditioned inpainting to place a given subject adhering to the scene affordance within the \textit{semantic masks}. For learning semantic masks, we leverage rich object-scene priors learned from the text-to-image generative models and optimize a novel parameterization of the semantic mask, eliminating the need for large-scale training. To the best of our knowledge, we are the first ones to provide an effective solution for realistic human placements in diverse real-world scenes. The proposed method can generate highly realistic scene compositions while preserving the background and subject identity. Further, we present results for several downstream tasks - scene hallucination from a single or multiple generated persons and text-based attribute editing. With extensive comparisons against strong baselines, we show the superiority of our method in realistic human placement.  
  \keywords{Spatial relations \and Human inpainting} 
\end{abstract}
\section{Introduction}
Given a background scene, humans can easily visualize how persons can interact with the scene in multiple ways. For, e.g., for a living room, one can imagine a person sitting on the sofa, or walking towards the door. To understand this relationship between humans and scenes, J.J. Gibson coined the term \textit{affordances}~\cite{gibson1978ecological} which points to the interaction between objects and the environment. Designing computational models for this task is extremely challenging and is crucial for common sense visual understanding. Earlier methods for human affordance predictions are constrained by the specific datasets~\cite{direct-affordance1,direct-affordanc2-binge,direct-affordance3}. To generalize affordance predictions, a recent method~\cite{kulal2023putting} trained with a large dataset of human videos to place humans in a given bounding box. However, this formulation only models the \textit{local} human affordance in the given bounding box but cannot reason about the \textit{global} human affordance, such as where a person can sit or stand. In this work, we aim to learn generalized local and global human affordances \textit{without needing large-scale training}. Specifically, we guide the affordances through action prompts (e.g., \textit{`a person sitting on sofa'}). 

\begin{figure}[t]
    \vspace{-2mm}
    \centering
    \includegraphics[width=0.86\linewidth]{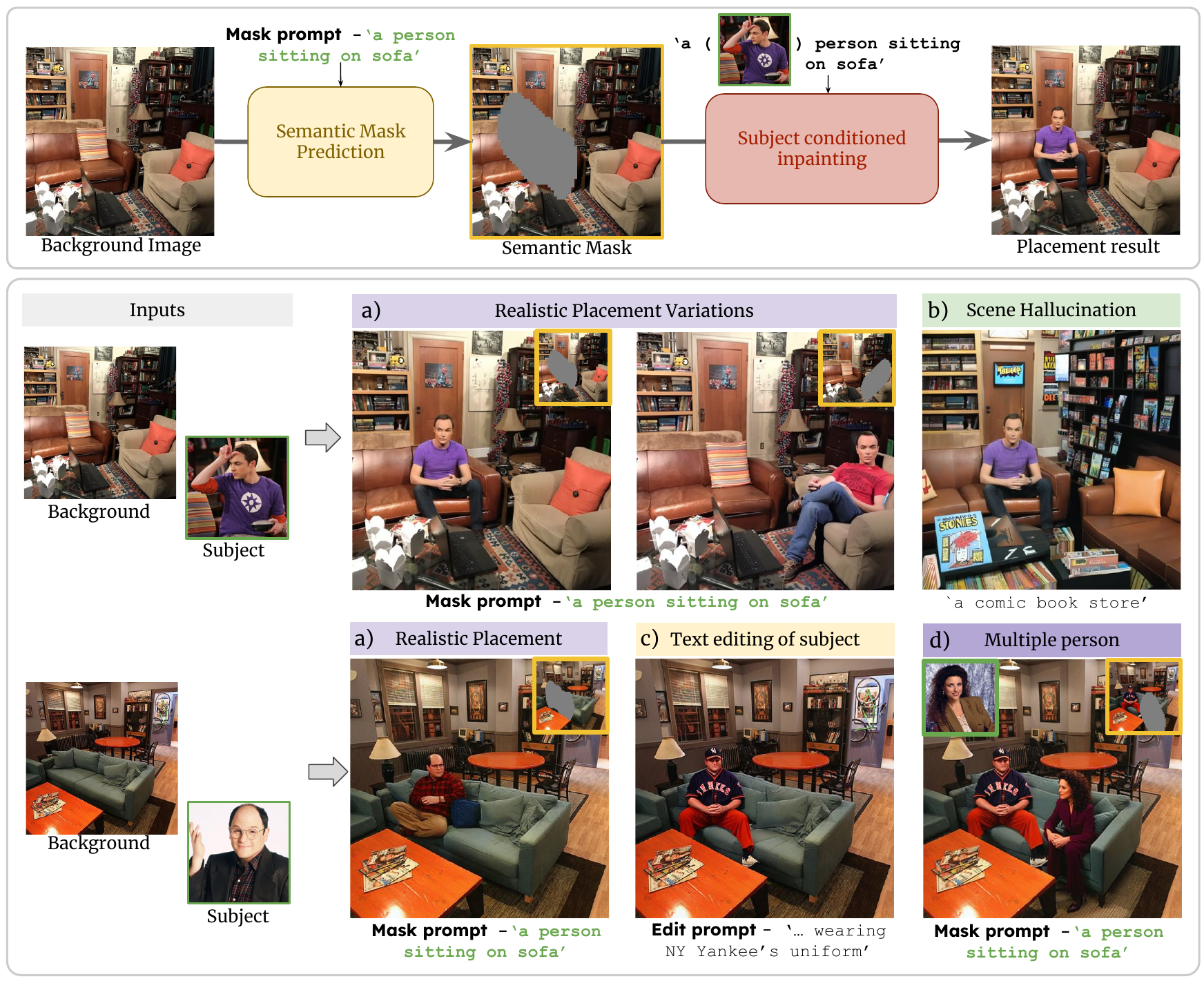}
    \vspace{-2mm}
    \caption{\textbf{(Top)}: Proposed approach for text-based placement of humans. Given a background image, we predict the plausible semantic region compatible with the text prompt to place humans. Next, given a few subject images, we perform subject-conditioned inpainting to realistically place humans in appropriate poses following the scene affordances. \textbf{(Bottom)}: Our method enables \textbf{a)} realistic human placements at diverse locations and poses and several downstream applications \textbf{b)} scene hallucination by generating compatible scenes for the given pose of the human \textbf{c)} text-based editing of the human and \textbf{d)} placing multiple persons.}
    \vspace{-8mm}
    \label{fig:teaser}
\end{figure} 

We model the human affordance of a scene as \textit{semantic masks} covering plausible image regions to place humans. Training a model to predict the \textit{semantic masks} for \textit{unseen} objects (e.g. person) is challenging primarily due to the unavailability of datasets. Typical image description datasets have annotations for what is present in the scene, such as object locations or scene captions, and do not provide information regarding the \textit{affordances}. Existing approaches~\cite{kulal2023putting,seeing-the-unseen} for affordance prediction adapt these datasets by inpainting objects to create synthetic datasets of image pairs \textit{with/without} objects. However, these methods are extremely expensive to train and are limited to the objects on which the model is trained.
To facilitate a more flexible setting, we build upon large text-to-image~\cite{ldm} (T2I) generative models. These models have rich priors for object-scene composition; which are implicit in the generation process. \textit{Can we use these priors from T2I models to obtain text-guided affordances for a given scene?} To this end, we propose an insightful approach that uses score distillation sampling (SDS)~\cite{sds} from T2I models to optimize  \textit{semantic masks} for learning human affordances. Specifically, given a background image and an action prompt (e.g., \textit{`a person sitting on the sofa'}), we use SDS loss to optimize parametric \textit{semantic mask} to localize regions to place humans. Naively parameterizing \textit{semantic masks} in the pixel space leads to collapse where the optimized mask covers the full image. To regularize the semantic masks, we propose a novel blob-based parameterization that constrains the semantic mask to a suitable region according to the required human pose.  

The obtained \textit{semantic mask} is used to perform subject-conditioned inpainting for scene completion. Given a few images of the subject, we project it into the text token embedding space of pretrained T2I model~\cite{ldm} using Textual Inversion~\cite{textual-inversion}. Next, we use the learned token embedding and the action prompt to place the subject using the inpainting pipeline~\cite{ldm}. Large T2I models have excellent inpainting capability and can adjust the person's pose given a \textit{semantic mask} for realistic scene completions (Fig.~\ref{fig:teaser}). We call this problem for placing persons following affordances as \textbf{Semantic Human Placement (SHP)}. To the best of our knowledge, we provide the first solution to solve SHP for realistic human placement in in-the-wild scenes. 


We present extensive results for SHP on diverse indoor and outdoor scenes. Further, we show multiple downstream applications of \textit{scene hallucination} conditioned on the single or multiple generated persons. Subject conditioned inpainting with T2I models enables text-based editing of the generated person. Additionally, our method generalizes to place diverse objects beyond humans in the scene. We compare our method against several baselines, including recent Vision Language Models models ~\cite{achiam2023gpt,llava}, and achieve superior scene completion with background preservation evident in quantitative evaluations and user study. In summary, our primary contributions are:

\begin{enumerate}
    \item A novel problem formulation of \textbf{Semantic Human Placement (SHP) for} realistically placing a given subject in background scenes.
    \item Method to effectively parametrize semantic masks and learning masks using distillation from text-to-image models without large-scale training. 
    \item A subject conditioned inpainting pipeline to generate identity-preserving realistic human placements from a few subject images.
    \item Demonstrate the efficacy of our models on downstream tasks of person and scene hallucination, composing multiple persons and placing other objects.
\end{enumerate}
\section{Related Work}
\textbf{Object and scene affordances.} Inpired by the earlist notion of afforances~\cite{gibson1978ecological}, several works have been proposed to perform affordance prediction~\cite{chuang2018learning , delaitre2012scene,quadflieg2017neuroscience,fouhey2015defense,grabner2011makes,gupta20113d,jiang2013hallucinated,li2019putting,wang2017binge}. These works have focused on modeling human-object~\cite{cao2021reconstructing,gkioxari2018detecting,koppula2013learning,yao2010modeling, zhu2014reasoning} and human-scene affordances~\cite{wang2021synthesizing,lee2002interactive,cao2020long}. An interesting framework is to learn human affordances from videos~\cite{quadflieg2017neuroscience,fouhey2015defense,wang2017binge}. In an insightful work~\cite{direct-affordanc2-binge} learn human affordances from a large video dataset of sitcom shows. Specifically, they learn plausible human poses for each of the sitcom scenes, which limits their applicability in learning the affordances of diverse real scenes. Brooks~\cite{brooks2022hallucinating} train a generative model to learn pose-conditioned human and scene generation to generate realistic compositions. Further, they allow for diverse scene generations that are compatible with a given human pose. More recently, Kulal ~\cite{kulal2023putting} trained a large diffusion model to place humans in a defined bounding box with appropriate local affordances. For learning object-scene, Ramrakhya~\cite{seeing-the-unseen} curated a large-scale dataset of image pairs with/without the object of interest to train object affordances for a given scene. A concurrent work SmartMask~\cite{singh2024smartmask} performs a large-scale training of diffusion model to predict object masks. All the above methods, although learning good affordances, either rely on large-scale training, require guidance as a bounding box, or do not generalize to novel object categories. Our work aims to learn local and global human affordances in diverse scenes without a need for large-scale training and also to support affordance for novel objects.

\noindent \textbf{Inpainting.} The task for inpainting is to fill a given masked region with plausible pixels to facilitate realistic image completion. Earlier attempts for inpainting explore local image features~\cite{old_inp1,old_inp2,old_inp3,old_inp4}. Later, several works were designed to perform inpainting with large-scale training of CNNs~\cite{cnn_inp1,cnn_inp2,cnn_inp3,cnn_inp4,cnn_inp5} and transformers~\cite{trans_inp1,trans_inp2,trans_inp3}. Recent works leverage the exceptional generation quality of diffusion models to perform inpainting~\cite{diff_inp1,paint-by-example,diff_inp2}. The closest to our work is guided inpainting by leveraging rich priors learned in large text-to-image diffusion models~\cite{ldm}. A seminal work~\cite{paint-by-example} performs a reference image-guided inpainting instead of using the text prompts. Later, several works were proposed that advanced the idea of reference-guided inpainting with single~\cite{single_image_guide,single_image_guide_inp,zhang2023paste} or multiple subject images~\cite{multi_image_guide_inp} when given with an inpainting mask. In this work, we condition the inpainting with text along with the reference image without needing an inpainting mask.  

\noindent \textbf{Diffusion Models and personalization.} Diffusion models have become state-of-the-art for image generation and enable several downstream editing applications~\cite{diffusion-beats-gan,ddim,ddpm,prompt2prompt,diffusion-im2im}. These models scale extremely well when trained with large-scale image-text pairs for conditional generation~\cite{ldm,dalle,dalle-2,imagen}.
In contrast to GAN-based editing methods~\cite{shen2020interpreting,parihar2022everything,motionstyle,park2020swapping,isola2017image}, which are limited to the data domains they are trained on, text-to-image models enable exceptional editing capabilities for in-the-wild real images~\cite{meng2021sdedit,imagic,zeroshotimg2img} and enable various controls in the generation process ~\cite{prompt2prompt,attend-excite,zhang2023adding_controlnet,tumanyan2022plugandplay}. The text-to-image models can be personalized for a given subject with few input images. In this line, popular works like Dreambooth~\cite{ruiz2023dreambooth}, Custom Diffusion~\cite{custom-diffusion} and Textual-Inversion~\cite{textual-inversion} have proven effective in embedding concepts within text-to-image models with test-time tuning. Following this, there are several works~\cite{chen2023anydoor,e4t,profusion,celeb-basis,Ye2023IPAdapterTC,Wang2024InstantIDZI} that aim to improve personalization in terms of training efficiency and identity preservation. For simplicity, we use Textual inversion \cite{textual-inversion} for learning a given subject for subject-conditioned inpainting. To distill the rich representations learned from text-to-image models to other domains, Score Distillation Sampling (SDS) is proposed. In the seminal work on dreamfusion~\cite{sds}, SDS loss is used to optimize NeRF~\cite{wang2021nerfmm} for text-to-3D generation. Later, several modifications to the original SDS loss are proposed~\cite{sds,noisefreeSDS,dds} to improve high-frequency details. As our goal is to optimize semantic mask instead of generation, we use SDS loss as it works well in capturing low-level features.


\section{Method}
\begin{figure}[t]
    \centering
    \vspace{-4mm}
    \includegraphics[width=1.0\linewidth]{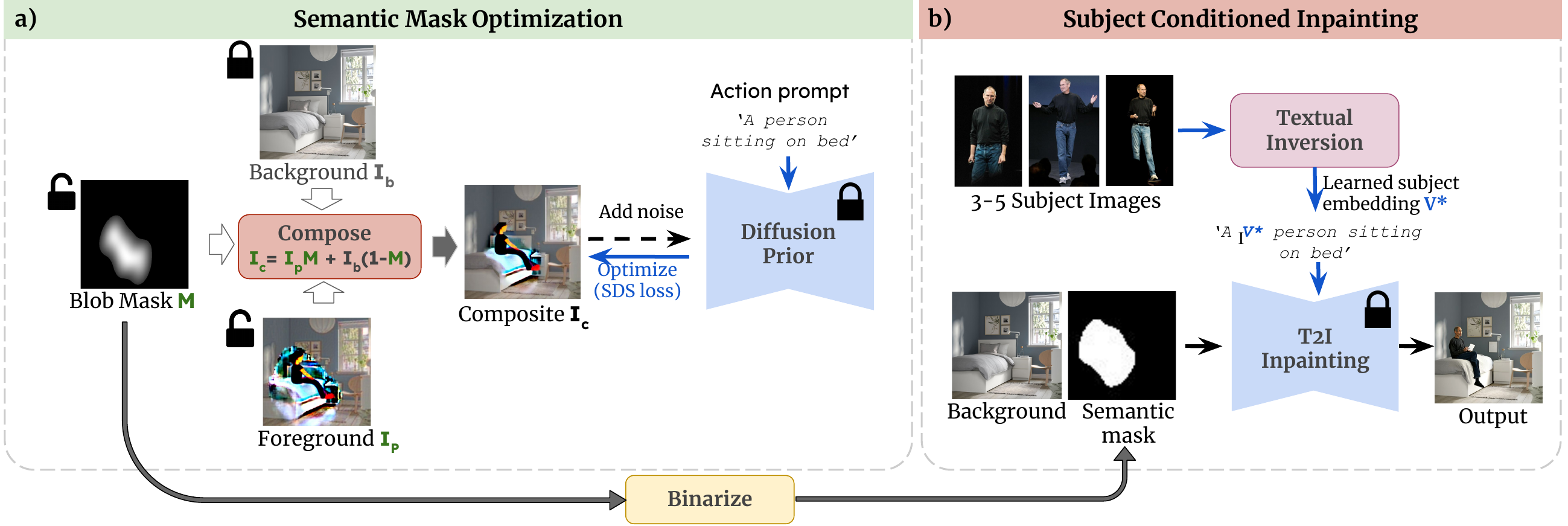}
    \vspace{-6mm}
    \caption{Our approach consists of two stages: \textbf{a) Semantic Mask Optimization.} Given a background image $\mathcal{I}_b$, we initialize a blob mask $\mathcal{M}$ parameterized as Gaussian blobs and a foreground person image $\mathcal{I}_p$. These two images are combined to form a composite image $\mathcal{I}_c$, which is used to compute SDS loss with the action prompt. During optimization, only $\mathcal{M}$ and $\mathcal{I}_p$ are getting updated via $\mathcal{I}_c$. After training $\mathcal{M}$ converge to a plausible human placement region, which is then used for inpainting. \textbf{b) Subject conditioned inpainting.} Given a few subject images, we perform Textual Inversion to obtain its token embedding $\mathbf{V*}$. Next, we use the inpainting pipeline of T2I models to perform personalized inpainting of the subject.} 
    \vspace{-6mm}
    \label{fig:method-combined-image}
\end{figure}

The goal of \textbf{Semantic Human Placement (SHP)} is to predict human affordances in a given background scene to place a given subject realistically. Our approach for SHP involves two stages, as shown in Fig.~\ref{fig:method-combined-image}. Firstly, we learn human affordances defined as \textit{semantic mask} in the background image with text guidance locating the regions where a person can be placed plausibly (Sec.~\ref{subsec:mask-gen}). Next, we perform subject-conditioned inpainting using the \textit{semantic mask} and a few subject images for a realistic scene composition (Sec.~\ref{subsec:cond-inpaint}).

\subsection{Semantic mask generation} 
\label{subsec:mask-gen}

Predicting semantic masks for objects (humans) that are not present in the scene is extremely challenging. The popular image description datasets only describe the objects present in the image. Existing methods~\cite{seeing-the-unseen,li2019putting} adapt these datasets by inpainting objects to create a paired dataset for object affordance and perform large-scale training to predict affordance. We take a radical approach to learning affordance using knowledge from T2I~\cite{ldm} models, \textit{eliminating the need for data curation and expensive training}. T2I models have excellent object-scene compositional reasoning implicit in the generation process. An effective way to distill this knowledge is to use Score Distillation Sampling (SDS) loss~\cite{sds}. SDS and its variations have proven extremely effective for realistic text-to-3D generation by optimizing underlying 3D representation~\cite{sds,tang2023dreamgaussian,Chen2023Textto3DUG}. 

The success of SDS loss raises a natural question: \textit{Can we use the SDS loss to learn the semantic mask for object placements?} If yes, \textit{what could be an effective semantic mask representation for optimization?} Directly learning the semantic mask in the pixel space could easily lead to collapse covering all the image pixels (supplementary). We propose a novel \textit{semantic mask} representation, parameterized as a set of interconnected Gaussian blobs. The proposed representation is extremely efficient, with only a few learnable parameters and fast rendering. Furthermore, it's highly compatible with learning through SDS loss.


\subsubsection{Training.}
\begin{wrapfigure}{r}{0.55\textwidth}
    \centering
    \vspace{-8mm} 
    \includegraphics[width=\linewidth]{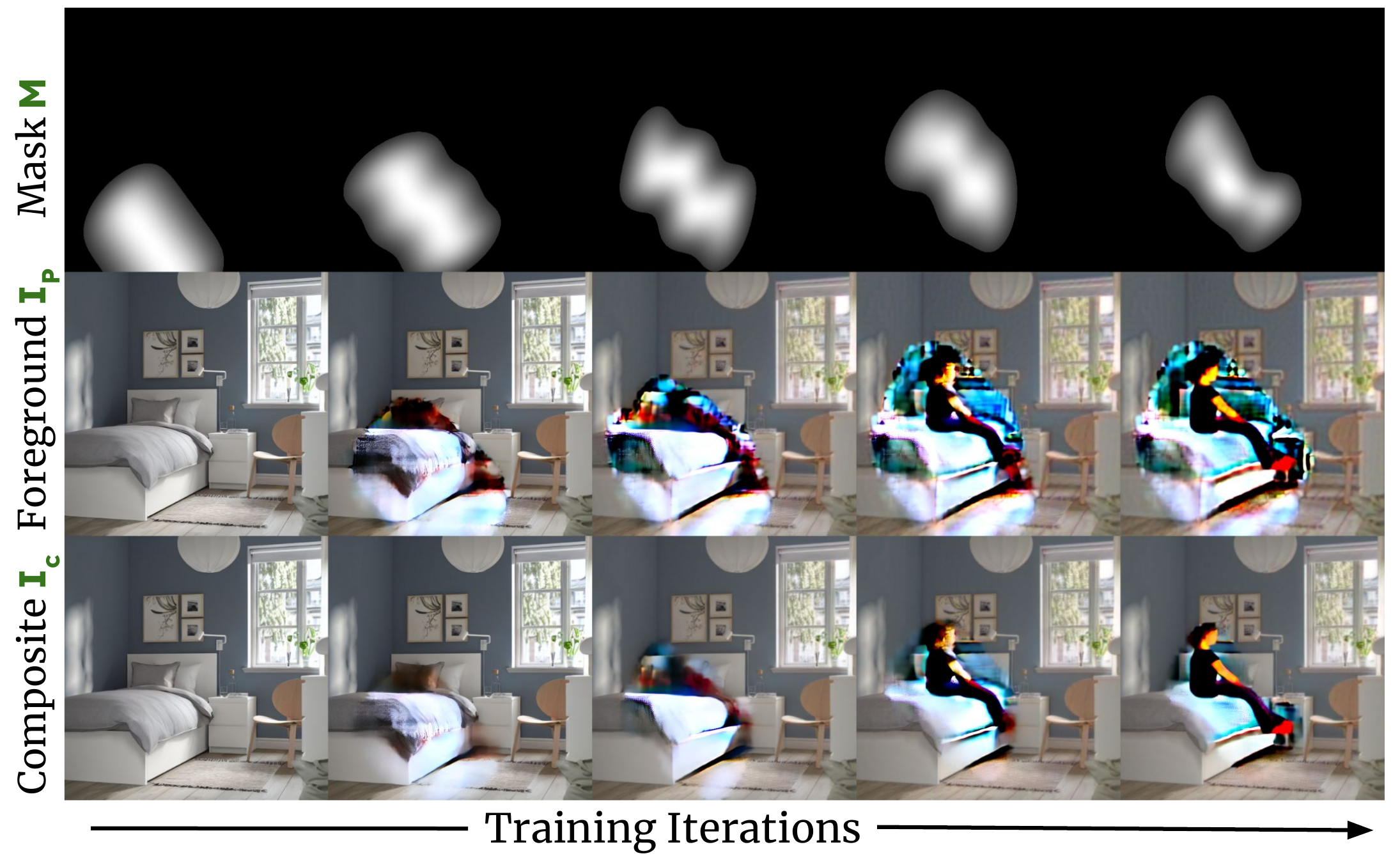}
    \vspace{-8mm} 
    \caption{\textbf{Training progression,} of semantic mask $\mathcal{M}$ and person image $\mathcal{I}_p$ with SDS loss with prompt \textit{`A person sitting on bed'}. $\mathcal{M}$ starts to converge at the appropriate person region following prompt.}
    \vspace{-9mm} 
    \label{fig:mask-optimization-phase1}
\end{wrapfigure}

For effectively learning the parameterized \textit{semantic mask} $\mathcal{M}$ on a background image $\mathcal{I}_b$, we create a framework involving a learnable foreground person image $\mathcal{I}_p$ initialized as a copy of $\mathcal{I}_b$. The major function of the $\mathcal{I}_p$ is to aid the learning of \textit{semantic mask} parameters. Specifically, at each training iteration, we combine $\mathcal{I}_p$ and $\mathcal{I}_b$ using mask $\mathcal{M}$ to obtain a combined image $\mathcal{I}_c$ which will be used to compute SDS loss. The intuition is that as the training progresses, $\mathcal{I}_p$ will generate a person, and $\mathcal{M}$ will converge to the person's location and shape. This will, in turn, generate a person in $\mathcal{I}_c$ after composition, as shown in Fig.~\ref{fig:mask-optimization-phase1}. This formulation enables the learning of semantic plausible regions for human placement in $\mathcal{M}$. To compute SDS loss, a noisy version of $\mathcal{I}_c$ is passed through

\noindent the T2I model with an action prompt (e.g.,\textit{`a person sitting on a bed'}). The gradients of the loss are computed with respect to $\mathcal{I}_c$ and are further backpropagated to compute gradients and update the image $\mathcal{I}_p$ and parameters of $\mathcal{M}$. Once trained, we binarize the optimized \textit{semantic mask} $\mathcal{M}$ and use it for subject-conditioned inpainting on the background image. We note that the optimized mask is not pixel-perfect due to the Gaussian blob parameterization. However, the downstream inpainting based on T2I models expects a coarse semantic mask to allow room for adapting the foreground during inpainting Fig.~\ref{fig:mask-sensitivity-ablate}. Further, this enables us to generate pose variations of the subjects and text-based editing of the placed person, which is not possible with a pixel-perfect mask Fig.~\ref{fig:inpaint-downstream}.

\subsubsection{Mask parametrization.} 
\begin{wrapfigure}{r}{0.5\textwidth}
    \vspace{-8mm} 
    \centering
    \includegraphics[width=1.0\linewidth]{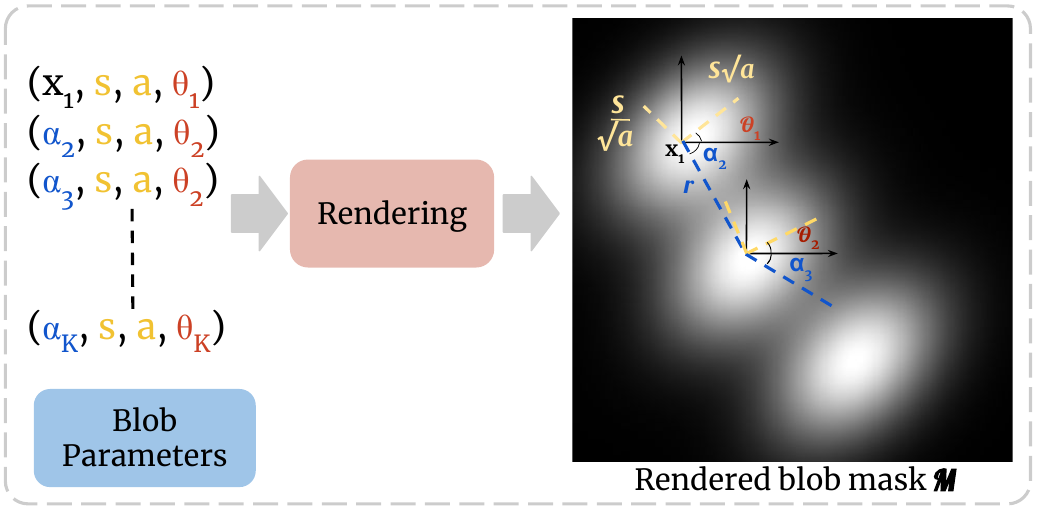}
    \vspace{-8mm}  
    \caption{\textbf{Blob mask parameterization.}}
    \label{fig:blob-parameterization}
    \vspace{-10mm}  
\end{wrapfigure}

For effectively learning of \textit{semantic mask} using SDS loss, we parameterize $\mathcal{M}$ as a set of $\mathbf{K}$ interconnected Gaussian blobs (Fig.~\ref{fig:blob-parameterization}). Following ~\cite{epstein2022blobgan}, we use simple ellipsoid representation for blobs by defining center location $\mathbf{x} \in [0,1]^2$, scale $\mathbf{s} \in \mathbb{R}$, 
aspect ratio $\mathbf{a} \in \mathbb{R}$, rotation angle $\theta \in [-\pi/2,\pi/2]$. We tie the blobs by keeping a fixed distance $\mathbf{r} \in \mathbb{R}$ between 

\noindent the center of consecutive blobs. Precisely, the center location $\mathbf{x}_i$ for the $i^{th}$ ($i>1$) blob is computed using the previous blob center $\mathbf{x}_{i-1}$ using Eq.~\ref{eq:eq1}.

\begin{equation}
\mathbf{x}_{i} =  \textbf{x}_{i-1} + \begin{bmatrix} \mathbf{r} \cdot \cos(\alpha_{i}) \\ \mathbf{r} \cdot \sin(\alpha_{i}) \end{bmatrix}
\label{eq:eq1}
\end{equation}

\noindent where $\alpha_i \in \mathbb{R}$ is the relative angle between the blob centers from $\mathbf{x}$-axis. Finally, the total blob parameters defining \textit{semantic mask} are the following: 

\begin{equation}
    \gamma_1 \equiv (x_1, s, a, \theta_1) \;\;\; and \;\;\; \gamma_i \equiv (\alpha_i, s, a, \theta_i) \;\; for \; i \in \{2,K\} 
\end{equation}  

\noindent Next, we use parameters $\gamma_i$ to create a mask image $\mathcal{M}_i \in [0,1]^{H \times W}$ for $i^{th}$ blob. We perform this operation in a differentiable manner to enable optimization of the parameters. Specifically, we compute the Mahanobolis distance between every pixel coordinate location $\mathbf{x}_{grid}$ from the $i^{th}$ blob center ($\mathbf{x}_i$) to assign the intensity value at $\mathbf{x}_{grid}$

\begin{equation}
\centering 
{D^m}(\mathbf{x}_{\text{grid}}, \mathbf{x}_i) = (\mathbf{x}_{\text{grid}} - \mathbf{x}_i)^T (R \cdot \Sigma \cdot R^T)^{-1} (\mathbf{x}_{\text{grid}} - \mathbf{x}_i)
\end{equation}
\vspace{-6mm}
\begin{equation}
\mathcal{M}_i[\mathbf{x}_{grid}] = \exp(-0.5 * {D^m}(\mathbf{x}_{grid}, \mathbf{x}_i))
\end{equation}  

\noindent where $\Sigma = c.\begin{bmatrix} {\delta_x}^2 & 0 \\ 0 & {\delta_y}^2 \end{bmatrix}$, R is the rotation matrix based on $\theta_i$ , $c=0.02$ controls the sharpness of each blob and ${\delta_x} = s/\sqrt{a}$, and ${\delta_y} = s \sqrt{a}$. Finally, we obtain aggregated \textit{semantic mask} $\mathcal{M}$ by taking the pixel-wise mean of the individual blob masks $\mathcal{M}_i$. During training, we only optimize the first blob center location $\mathbf{x}_1$, all rotation angles $\theta_i$, and relative angles $\alpha_i$. This provides enough flexibility for the mask to reach the appropriate location and adjust based on the required human pose. Empirically, we have observed fixing $\mathbf{s}$, $\mathbf{a}$, and $\mathbf{r}$ to a constant for all the blobs results in better convergence of the mask (suppl. material). 

\subsection{Subject conditioned inpainting}  
\label{subsec:cond-inpaint}
Given a few unposed images of a subject, we aim to place them in the background image following affordances using the optimized \textit{semantic mask} $\mathcal{M}$. Inpainting humans is extremely hard, as the person's pose should be appropriately changed based on the location of the mask (e.g. if we are inpainting a person on a chair, a sitting pose is more suitable). Naively, following a reference-based inpainting~\cite{paint-by-example} does not consider these person-scene affordances and results in unnatural compositions (Fig.~\ref{fig:inpainting_pbe_comparison}). We leverage rich human-scene priors from the T2I~\cite{ldm} models to perform plausible inpainting of humans respecting the local affordances. The text-guided inpainting pipeline of T2I models generates plausible outputs adapting the person's pose according to the background. We ask: \textit{How can we adapt this inpainting pipeline for subject conditioned inpainting?} We propose to pass the subject-specific knowledge in T2I through the input text conditioning used for inpainting (Fig.~\ref{fig:method-combined-image}b)). Specifically, we learn a token embedding $\mathbf{V*}$ representing our subject from a few input images using Textual Inversion~\cite{textual-inversion}. Next, we use $\mathbf{V*}$ along with the inpainting prompt (e.g., \textit{`A} $\mathbf{V*}$ {person sitting on a sofa'}) to condition the inpainting pipeline of T2I model. This simplistic framework for conditioning generates realistic human placements that follow the scene affordances, capitalizing on the rich object-scene priors from the T2I models (Fig.~\ref{fig:method-combined-image}). Further, our inpainting can benefit from improved text-based personalization methods in T2I models. 


\subsubsection{How accurate should the semantic mask be?} 
\begin{wrapfigure}{r}{0.5\textwidth}
    \centering
    \vspace{-9mm}
    \includegraphics[width=1.0\linewidth]{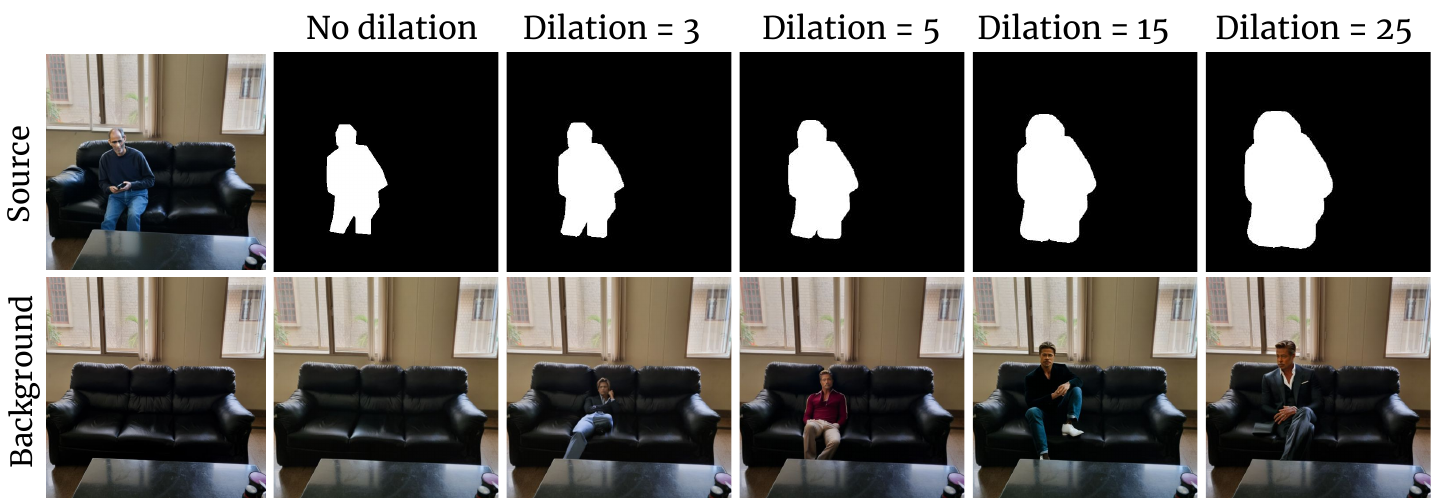}
    \vspace{-7mm}
    \caption{\textbf{Pixel-perfect mask is not effective for T2I inpainting.}}
    \vspace{-8mm}  
    \label{fig:mask-sensitivity-ablate}
\end{wrapfigure}

We analyze the sensitivity of our inpainting pipeline on the preciseness of \textit{semantic mask} $\mathcal{M}$ in Fig.~\ref{fig:mask-sensitivity-ablate}. Given a background image, we first run our method to place a person in the scene to obtain source image. Next, we compute the segmentation mask of the person using SAM~\cite{sam}. The obtained person mask is compatible with the background and can be used for inpainting. We ablate over the coarseness of the mask by  

\noindent iteratively dilating the segmentation mask and performing inpainting. Using the original mask for inpainting is too stringent and does not place the person as the T2I inpainting pipeline is stochastic. On increasing the dilation, inpainting is successful but lead to background distortions. We conclude that using a pixel-perfect \textit{semantic mask} hurts inpainting and a coarse mask is needed for realistic inpainting results. This further motivates our blob parameterization which creates a coarse mask \textit{semantic mask} for placing humans.

\section{Experiments}
We perform extensive experiments to evaluate our method and provide detailed ablations. We use Stable Diffusion models as our representative T2I models in all the experiments. Given a background image, we perform test time optimization to obtain the semantic mask for human placement. Specifically, we optimize the blob parameters for the mask for $1000$ iterations using SDS loss~\cite{sds} with a guidance scale of $200$. Next, we perform subject-conditioned inpainting by performing textual inversion with $3-5$ subject images. Further details about implementation are provided in the supplementary. In this section, we first discuss datasets and metrics (Sec.~\ref{subsec:data_metrics}) followed by experiments on placing persons (Sec.~\ref{subsec:placing_person}), downstream applications (Sec.~\ref{subsec:down-appl}) and finally ablation studies (Sec.~\ref{subsec:ablations}). 

\begin{figure}[t] 
    \vspace{-2mm}
    \centering
    \includegraphics[width=0.9\linewidth]{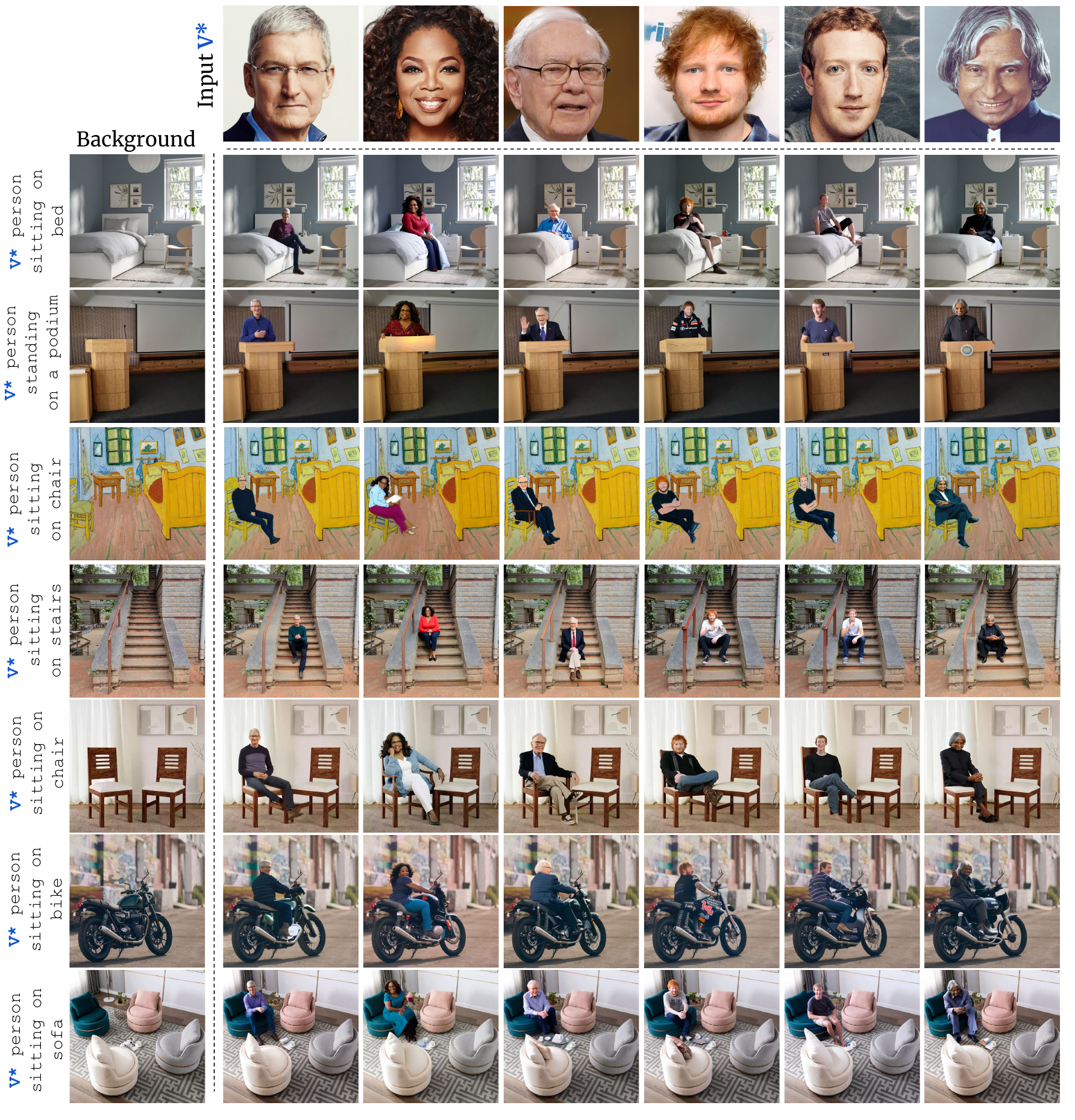}
    \vspace{-4mm}
    \caption{Realistic human placement in diverse indoor and outdoor scenes.}
    \label{fig:ours-qual-results}
    \vspace{-8mm}
\end{figure} 

\subsection{Dataset and metrics}
\label{subsec:data_metrics}

\noindent \textbf{Dataset.} For evaluation, we collected a dataset consisting of $30$ in-the-wild indoor and outdoor background images from the web, along with a few manually captured images in a university. For personalization, we select $15$ celebrity images, including scientists, actors, sports persons, and tech executives, and collect $4-5$ images for each subject. For placement, we use action-based text prompts such as \textit{`a person sitting on sofa'}, based on the background image. Additional details about the prompts are provided in the supplementary. 

\noindent \textbf{Metrics.} We evaluate realistic human placement on the following three aspects: 

\noindent \textbf{a) Text Alignment} - The generated image after placing a human should follow the given action prompt for desired text-based placement. We compute the cosine similarity between the CLIP~\cite{clip} image embedding of the generated image and CLIP text embedding of the provided prompt. Higher CLIP similarity (\textbf{CLIP-sim}) signifies that human placement follows the given prompt. 

\noindent \textbf{b) Person generation} - Inpainting model fails to generate a person if the inpainting mask is incorrectly placed (e.g., a person sitting on a window). To quantify correct person generation during inpainting, we detect humans in the generated images by applying SAM~\cite{sam} on the output and checking for an instance of the `person' class. We aggregate this metric as the percentage of images where a person is detected (\textbf{\% Person}), noting the accuracy of placement.

\noindent \textbf{c) Background Preservation} - To quantify the background preservation, we first segment out the human from the generated output using SAM~\cite{sam} and compute LPIPS~\cite{zhang2018unreasonable} in the \textit{non-human} regions between the background and inpainted image. For perfect placement, the background should be preserved during inpainting with lower LPIPS scores. 


We want to highlight that there is an inherent tradeoff between these metrics: Higher CLIP-sim and \% Person could be attained by using a large inpainting mask, but this results in significant background distortions (higher LPIPS). Hence, we aim to balance these metrics well for desired results. 

\subsection{Placing humans}
\label{subsec:placing_person} 
\begin{figure}[t]
    \centering
    \includegraphics[width=1.0\linewidth]{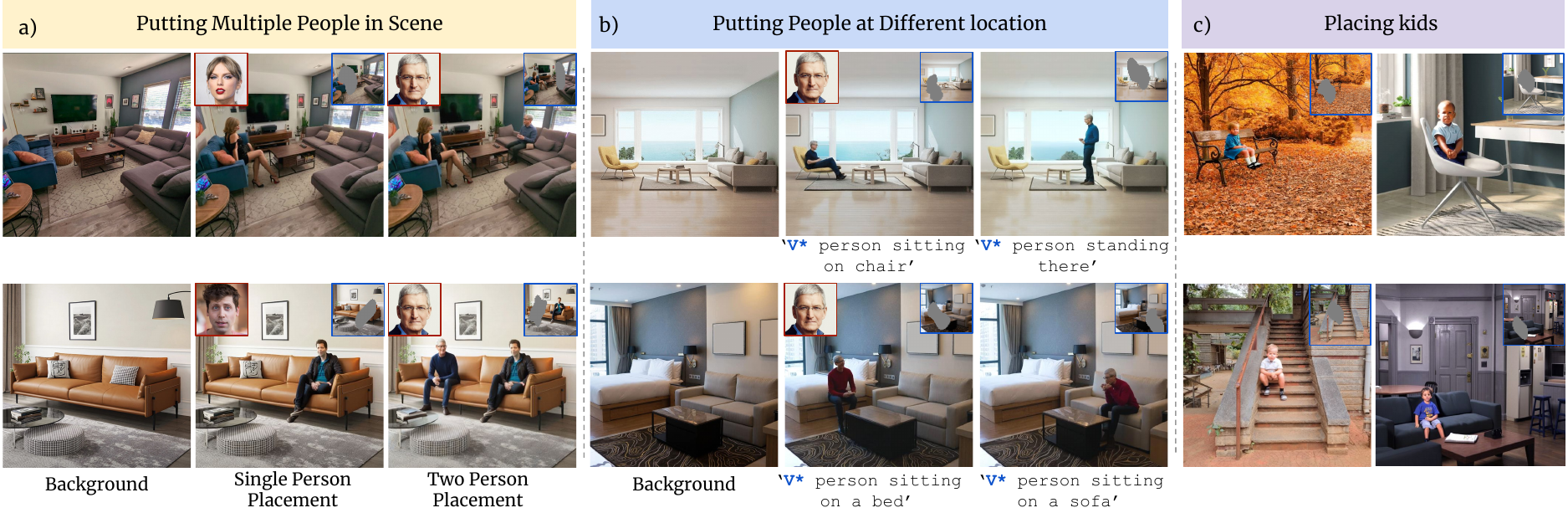}
    \vspace{-6mm}   
    \caption{Our method enables several interfaces for realistic human placement.} 
    \label{fig:multiple-people-location}
    \vspace{-8mm}  
\end{figure}

We present our qualitative results for Semantic Human Placement in Fig.~\ref{fig:ours-qual-results}. Our method generates highly realistic placement results while preserving the subject's identity and background. The obtained semantic masks properly localize the region in the background corresponding to the given action prompt. Also, observe the mask shape changes according to the desired human pose, which is compatible with the background (e.g., sitting pose in backgrounds $1$ and $3$ and bike riding pose in background $6$). Further, the proposed Text2Place enables diverse human placement interfaces as shown in Fig.~\ref{fig:multiple-people-location}. 

\noindent\textbf{Placing multiple persons sequentially} (Fig.~\ref{fig:multiple-people-location}a)) can be easily done by iteratively optimizing for two semantic masks. While optimizing for the semantic mask for the second subject, we add additional loss to minimize the overlap between the two masks. This helps generate disjointed semantic masks that are needed to place multiple persons. 

\begin{wraptable}{r}{0.50\linewidth}
    \vspace{-11mm}
    \centering
    \caption{Baseline comparison}
    \begin{adjustbox}{width=\linewidth}
        \begin{tabular}{c|c|c|c}
            \hline
            \textbf{Method}      & \textbf{LPIPS $\downarrow$} & \textbf{CLIP-sim $\uparrow$} & \textbf{\% Person $\uparrow$ } \\ 
            \hline
            GracoNet~\cite{graconet}   & 0.1090  & 0.2601 & 53.48 \\
            TopNet~\cite{topnet}   & 0.1162 & 0.2617  & 67.3  \\ 
            LLaVA~\cite{llava}               & 0.1296          & 0.2501               & 20.91            \\
            GPT4V~\cite{achiam2023gpt}       & 0.1059          & 0.2615               & 64.18            \\
            Ours (center)                    & \textbf{0.0845} & 0.2613               & 55.52            \\
            Ours                             & 0.0934 & \textbf{0.2726}      & \textbf{88.55}     \\
            \hline
        \end{tabular}
    \end{adjustbox}
    \vspace{-10mm}
    \label{table:comparison-baselines}
\end{wraptable} 

\noindent \textbf{Diverse placement with text guidance,} (Fig.~\ref{fig:multiple-people-location}b)) is achieved by providing different location prompts during semantic mask optimization.

\noindent \textbf{Placing kids.} Our notable blob mask representation enables the placement of smaller subjects by initializing with a relatively small blob size. This enables the realistic placement of kids in diverse scenes with background preservation. (Fig.~\ref{fig:multiple-people-location}c). 

\subsection{Comparisons}

\textbf{Human placement comparisons.} We compare our approach against the following baselines: Object placement baselines \textbf{1) TopNet}~\cite{zhu2023topnet} and \textbf{2) GracoNet}~\cite{graconet} trained on large-scale object placement dataset OPA~\cite{opa-dataset}. The OPA dataset contains significant human placement examples, making these baselines suitable for comparison. Additionally, we compare with state-of-the-art VLMs 

\begin{figure}[t] 
    \vspace{-4mm}
    \centering
    \includegraphics[width=0.9\linewidth]{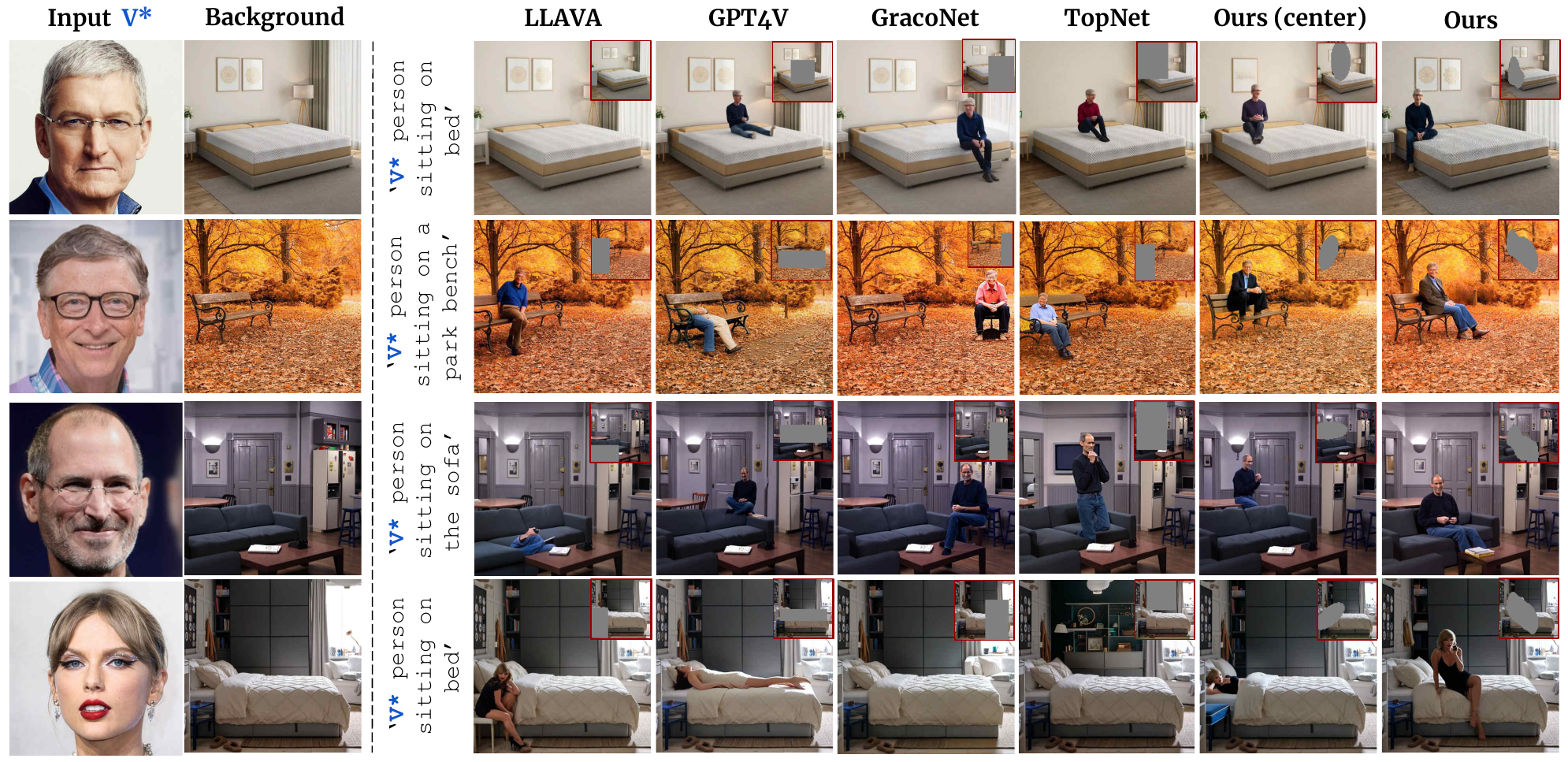}
    \vspace{-2mm}
    \caption{\textbf{Baseline Comparison:} Notably, LLaVa generates masks with semantically incorrect locations, while GPT4v produces excessively large mask sizes. GracoNet and TopNet predict bounding boxes at the correct location but of inaccurate size, whereas our method accurately determines the optimal location and size for person insertion.}
    \label{fig:comparison-baseline}
    \vspace{-6mm}  
\end{figure}  

\begin{wrapfigure}{r}{0.6\textwidth}
    \vspace{-8mm}
    \centering
    \includegraphics[width=\linewidth]{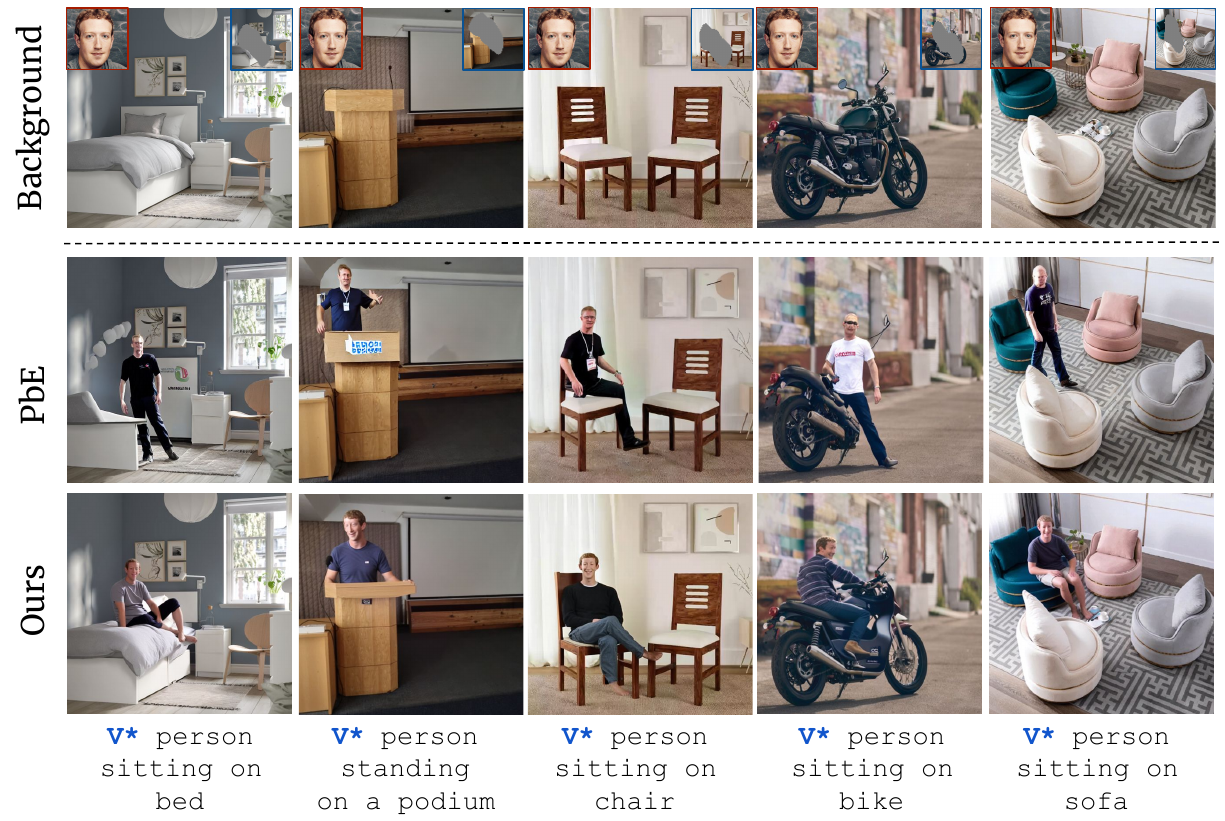}
    \vspace{-8mm}
    \caption{Inpainting comparison with PbE~\cite{paint-by-example}}
    \label{fig:inpainting_pbe_comparison}
    \vspace{-10mm} 
\end{wrapfigure} 

\noindent - \textbf{3) GPT4V~\cite{achiam2023gpt}} and \textbf{4) LLAVA}~\cite{llava}.
These models have excellent multimodal reasoning capabilities. For our experiments, we provide them with a background image and a prompt \textit{`predict a bounding box location for placing a human on the chair'}. 
From all the above baselines, the obtained bounding box coordinates are then used to create a semantic mask for inpainting in the next stage. To analyze the importance of the shape of the semantic mask vs the location, we compare it against baseline - \textbf{5) Ours (center)}, where we parameterize the semantic mask as a single blob with a learnable center. The scale $s$ is fixed to be $0.9$, and the orientation angle is randomly selected. The comparison results are present in Fig.~\ref{fig:comparison-baseline} and Tab.~\ref{table:comparison-baselines}. 

\noindent \textbf{Analysis.} GracoNet and TopNet are able to localize the bounding box correctly but struggle with generating the appropriate size of the box, resulting in unnatural placements. Likewise, in ours (center), the mask converges to the appropriate location for placement; however, it cannot place the person following local affordance due to the inappropriate shape of the semantic mask. This underscores the importance of semantic mask shape for forming realistic poses during placement. Bounding box masks generated by LLAVA and GPT4V are not able to cover the appropriate region to place a human. LLAVA mostly generates boxes on the left or bottom of the background image, resulting in no human placement or placement with unnatural poses. GPT4V generates masks at appropriate locations but fails to capture the shape needed to place humans. This results in unnatural human placement results. The same is evident from quantitative results, where GPT4V achieves the second-highest \%-Person score in comparison. Our method generates a person in most cases as both the semantic mask has an appropriate shape and location that follows the action prompts. Quantitatively, ours (center) achieves the best LPIPS score, suggesting the best background preservation; however, it has a lower CLIP-sim than ours, suggesting that the person is not appropriately placed following the text.


\begin{wrapfigure}{r}{0.55\textwidth}
    \vspace{-6mm}
    \centering
    \includegraphics[width=\linewidth]{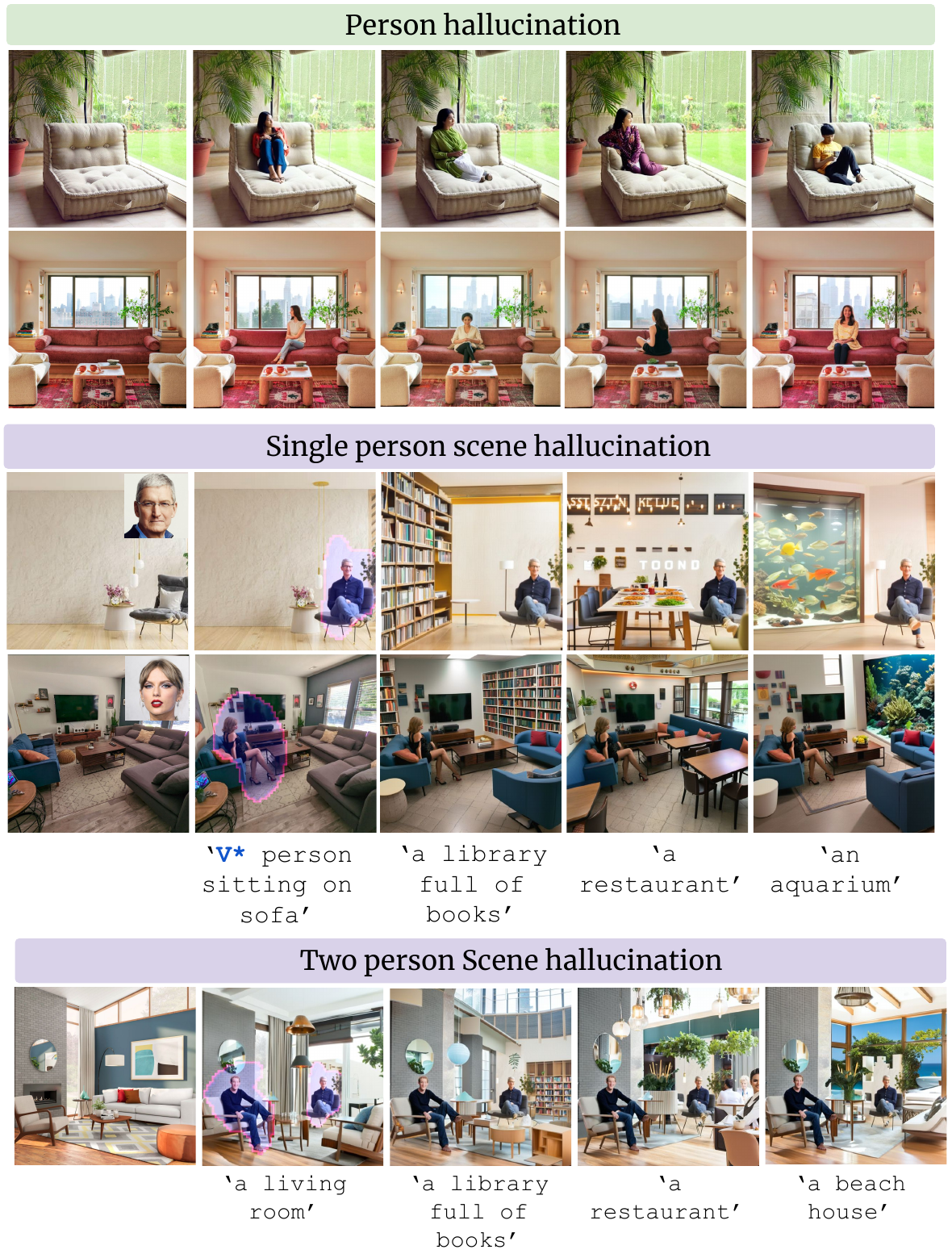}
    \caption{Person and scene hallucination}
    \label{fig:hallucination-results}
    \vspace{-14mm}
\end{wrapfigure} 
\noindent  

\noindent  

\noindent \textbf{Inpainting comparison.} We compare our inpainting method with a state-of-the-art reference-based inpainting method Paint-by-example 
(PbE) ~\cite{paint-by-example} in Fig.~\ref{fig:inpainting_pbe_comparison}. PbE is trained to inpaint a reference subject image given with an inpainting mask. For comparison, we use the semantic mask obtained from our method for both the inpainting methods. As PbE does not allow for text conditioning, it mostly generates a standing person. Moreover, the identity and the pose of the placed subject are inconsistent in general-purpose PbE.


\subsection{Applications}
\label{subsec:down-appl} 

\noindent \textbf{Person hallucination.} Given a background scene, we can hallucinate new persons by passing action prompts without subject conditioning (e.g., \textit{`a person sitting on sofa'}) (Fig.~\ref{fig:hallucination-results}). Our inpainting pipeline generates realistic outputs with humans in diverse poses consistent with the background, follows the text prompts, and preserves the subject's identity.

\noindent \textbf{Scene hallucination.} Another interesting application is to conditionally generate a scene compatible with the given pose of the person (Fig.~\ref{fig:hallucination-results}). To this end, we first place a human subject in the background using the predicted semantic mask. Next, we invert the semantic mask and perform outpainting of the region covered by the subject using the same T2I-based inpainting pipeline. We can conditionally generate diverse scenes by providing text prompts specifying the background (e.g., `\textit{a library full of books}').

\begin{wrapfigure}{r}{0.55\textwidth}
    \vspace{-8mm} 
    \centering
    \includegraphics[width=\linewidth]{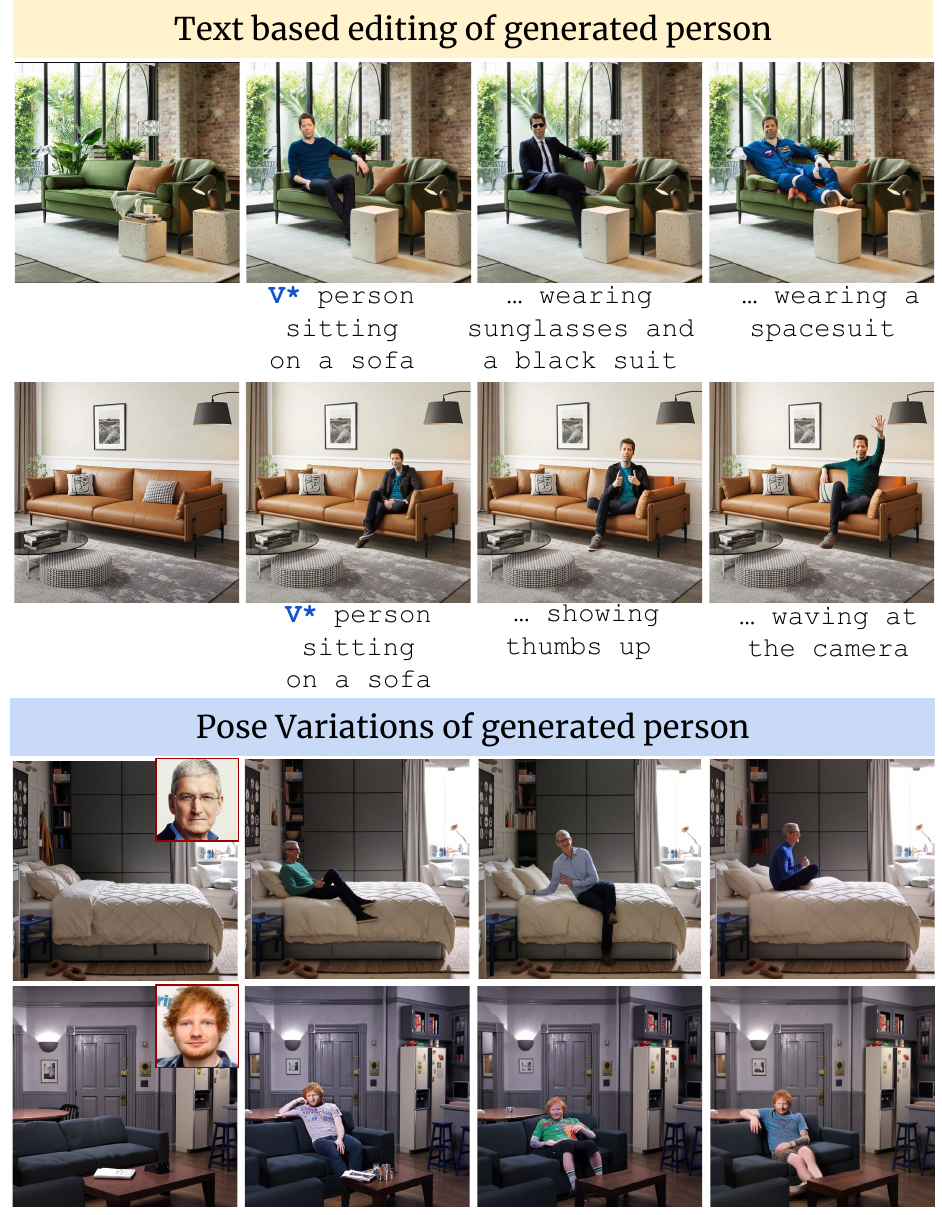}
    \vspace{-8mm}
    \caption{Applications of inpainting}
    \vspace{-6mm} 
    \label{fig:inpaint-downstream}
\end{wrapfigure}  

\noindent \textbf{Applications of subject conditioned inpainting.} The proposed pipeline for inpainting enables various downstream applications to control the generated image, as shown in Fig.~\ref{fig:inpaint-downstream}.

\noindent $\bullet$ \textit{Firstly}, we can perform text-based editing of the generated person by changing the text prompt for inpainting (e.g., \textit{`person -> person wearing a hat'}). As the inpainting is bound to modify only the semantic mask region, we achieve highly disentangled and localized editing of the generated subject.

\begin{wrapfigure}{r}{0.55\textwidth}
    \centering
    \vspace{-12mm}
    \includegraphics[width=\linewidth]{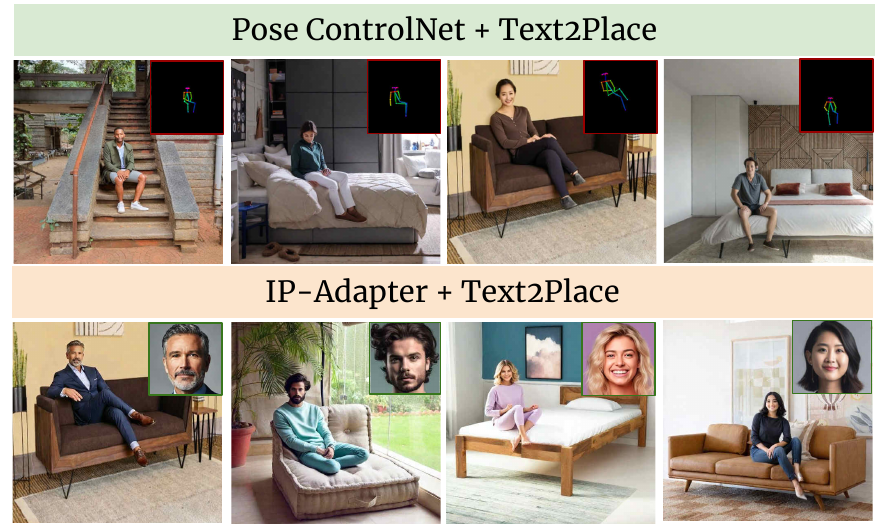} 
    \vspace{-8mm}
    \caption{Additional Conditioning}
    \label{fig:addn-condn}
    \vspace{-12mm}
\end{wrapfigure} 

\noindent $\bullet$ \textit{Secondly}, as we are using T2I diffusion model-based inpainting, which is inherently stochastic, we can generate diverse \textit{plausible} poses of the person from the same semantic mask and the action pro-mpt. Note that our mask 

\noindent
parameterization as a set of blobs allows us to generate diverse subject poses. 

\noindent \textbf{Additional Conditionings.} Our method can be integrated seamlessly with other conditioning approaches such as pose-conditioned ControlNet~\cite{zhang2023adding_controlnet} and IP-adapter
\noindent~\cite{ip-adap} for explicit conditioning on face identity (Fig.~\ref{fig:addn-condn}).  

\begin{wrapfigure}{r}{0.55\textwidth}
    \centering
    \vspace{-8mm}
    \includegraphics[width=\linewidth]{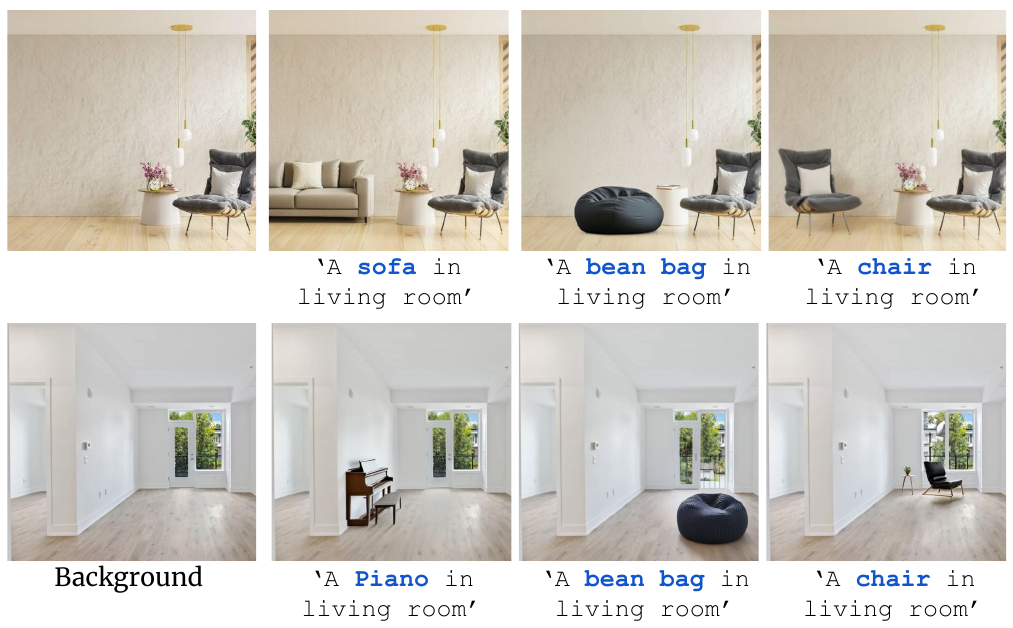}
    \vspace{-6mm}
    \caption{Placing objects beyond humans.}
    \label{fig:placing-different-objects}
    \vspace{-14mm}
\end{wrapfigure}

\noindent \textbf{Placing objects beyond humans.} perform these semantic edits requiring a \textit{loose} semantic 

\noindent mask, which would be challenging with a pixel-perfect semantic mask. Our method can easily be adapted to generate placements of various objects (Fig.~\ref{fig:placing-different-objects}). The obtained placements look highly realistic, with accurate placement of objects showing our approach's generalization without additional training. 

\subsection{Ablations} 
We ablate over the following blob parameters. Qualitative results for ablations are provided in the supplementary. 

\label{subsec:ablations}
\noindent \textbf{Scale of blob $(\mathbf{s})$.} is a crucial hyperparameter that controls the size of the blob; having a large $\mathbf{s}$ value results in significant background change, and a small $\mathbf{s}$ restricts the full body generation during the inpainting task. Empirically, we found $\mathbf{s}$ =0.6 gives the best results from Table~\ref{tab:ablations}a).

\begin{wraptable}{r}{0.45\textwidth}
\vspace{-10mm}
\caption{Ablation of blob parameters.}
\vspace{-4mm}
\begin{subtable}{\linewidth}
\centering
\caption{Ablation over blob scale $\mathbf{s}$}
\adjustbox{max width=0.9\textwidth}{
\begin{tabular}{c|c|c|c}
\hline
\textbf{Scale} & \textbf{LPIPS $\downarrow$} & \textbf{CLIP-sim $\uparrow$} & \textbf{\% Person $\uparrow$} \\ \hline
0.3            & 0.0537          & 0.2594              & 41.1                       \\
0.4            & 0.0806          & 0.2663              & 69.0                       \\
0.5            & 0.0858          & 0.2712              & 81.5                       \\
\textbf{0.6}   & \textbf{0.0904} & \textbf{0.2736}     & \textbf{90.6}              \\ 
0.7            & 0.1074          & 0.2729              & 96.0                       \\ \hline
\end{tabular}
}
\end{subtable}

\begin{subtable}{\linewidth}
\caption{Ablation over \# blobs}
\adjustbox{max width=1.0\textwidth}{
\begin{tabular}{c|c|c|c}
\hline
\textbf{\#blobs} & \textbf{LPIPS $\downarrow$} & \textbf{CLIP-sim $\uparrow$} & \textbf{\% Person $\uparrow$} \\ \hline
1                    & 0.1318               & 0.2780              & 93.0                       \\
3                    & 0.1305               & 0.2797              & 94.9                       \\
5                    & 0.0904               & 0.2736              & 90.6                       \\ 
7                    & 0.0780               & 0.2749              & 75.0                       \\ \hline
\end{tabular}
}
\end{subtable}
\label{tab:ablations} 
\vspace{-8mm}
\end{wraptable}

\noindent \textbf{Number of blobs.} Using a single blob 

\noindent doesn’t give freedom to generate the full body image of a person in different poses, so we use n number of blobs and fix the distance between each adjacent blob as discussed in Section 3.3, and the number n is a hyperparameter 
in our method, having fewer blobs restrict the mask diversity, but more blobs can cause bigger masks, leading to more background change. From Table~\ref{tab:ablations}b), we can see that n=5 gives a good balance between diversity and background change.

\section{Discussion}
\begin{wrapfigure}{r}{0.55\textwidth}
    \centering
    \vspace{-8mm}
    \includegraphics[width=1.0\linewidth]{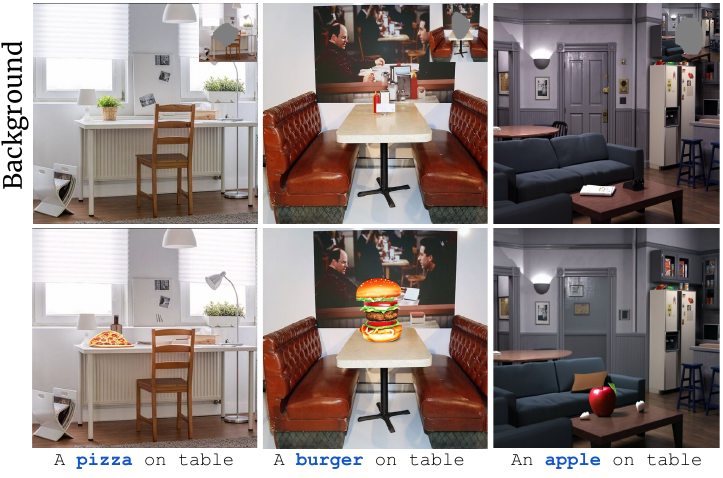}
    \vspace{-8mm}
    \caption{\textbf{Failure cases.}}
    \vspace{-8mm}  
    \label{fig:failure-case}
\end{wrapfigure}

\noindent \textbf{Limitations.}
While our approach excels in many aspects, it does have limitations (Fig.~\ref{fig:failure-case}. Notably, it may not effectively handle the placement of relatively 

\noindent small objects in a scene. This is because our blob mask representation occupies a significant image region. Using larger masks for inpainting results in changing the background or placing objects of inappropriate scale. Furthermore, our method lacks precise control over the generated person's poses, as tighter masks may hinder inpainting using T2I models shown in Fig.~\ref{fig:mask-sensitivity-ablate}. 

\noindent \textbf{Conclusion.} In this work, we propose the first method to learn text-conditioned human affordances for realistic human placement for in-the-wild scenes. We leverage rich priors learned in text-to-image (T2I) diffusion models to optimize plausible \textit{semantic masks} for human placement. The obtained masks are then used to perform subject-conditioned inpainting to place a given subject. Semantic masks are parameterized with a novel blob representation that adapts to the human pose during optimization with score distillation sampling in T2I models. The proposed method enables several downstream applications, including human and scene hallucination, placing multiple persons in a scene, and text-based editing of the generated subject. In summary, we propose a robust approach to perform text-guided placement of realistic humans in scenes without large pertaining.

\noindent \textbf{Ethics statement.} 
We acknowledge that our method can be used to generate images for malicious purposes, but similar to Stable Diffusion, the samples generated by our model can be watermarked. Since our model can be used to generate novel compositions for realistic human placement, it can be used for misleading generations. However, there’s a line of research for detecting fake samples from generative models, which we support. We believe the research contributions of this work outweigh the negative impacts. 

\noindent \textbf{Acknowledgements.} 
We thank Tejan Karmali for regular discussions and Abhijnya Bhat for reviewing the draft and providing helpful feedback. This work was partly supported by PMRF from Govt. of India (Rishubh Parihar) and Kotak IISc AI-ML Centre.

\clearpage 
\begin{center}
\Large \textbf{Appendix - Text2Place}
\end{center}  

\appendix 



\section{Implementation details}
\label{sec:supp-imple-details} 
We use Stable Diffusion models~\cite{ldm} as a representative T2I model in all of our experiments. We use Stable Diffusion-v1.4 to compute SDS loss (following ~\cite{poole2022dreamfusion}) for the semantic mask generation and Stable Diffusion-XL~\cite{ldm} for the subject conditioned inpainting task due to its superior inpainting capabilities. We perform instance-specific training for the background image to obtain a \textit{semantic mask}. Specifically, we optimize the blob parameters and the foreground image for $1000$ iterations using SDS loss~\cite{sds} with a guidance scale of $200$. The learning rate for the foreground image is $0.2$, and blob parameters $0.1$. We use an AdamW optimizer with no weight decay and a cosine scheduler. For other fixed blob parameters, we set the distance between blobs $r=0.01$, aspect ratio $a=2$, and blob scale $s=0.6$. Finally, we binarize the soft blob mask to obtain a semantic mask with a threshold value of $0.2$. We use standard hyperparameters from hugging face~\cite{roettger2021xltextualinversion,huggingface2022diffusers} for textual inversion use.

\section{Analysis on \textit{semantic mask} parameterization.}
In this section, we analyze the design choices for parameterization of the \textit{semantic mask} as blobs. We visualize the progression of the semantic mask, foreground image, and the combined image during training. \textbf{a)} \textbf{Pixel-wise mask}: We first try a naive baseline where the semantic mask is parameterized as a learnable gray-scale image. This simple parameterization results in a collapse of the mask image, resulting in the white pixels spread over the full image. This obtained mask is not useful for inpainting and cannot generate a person. \textit{semantic masks} resulting in natural placement outputs. 

\begin{figure}[h]
    \centering
    \vspace{-6mm}
    \includegraphics[width=0.85\linewidth]{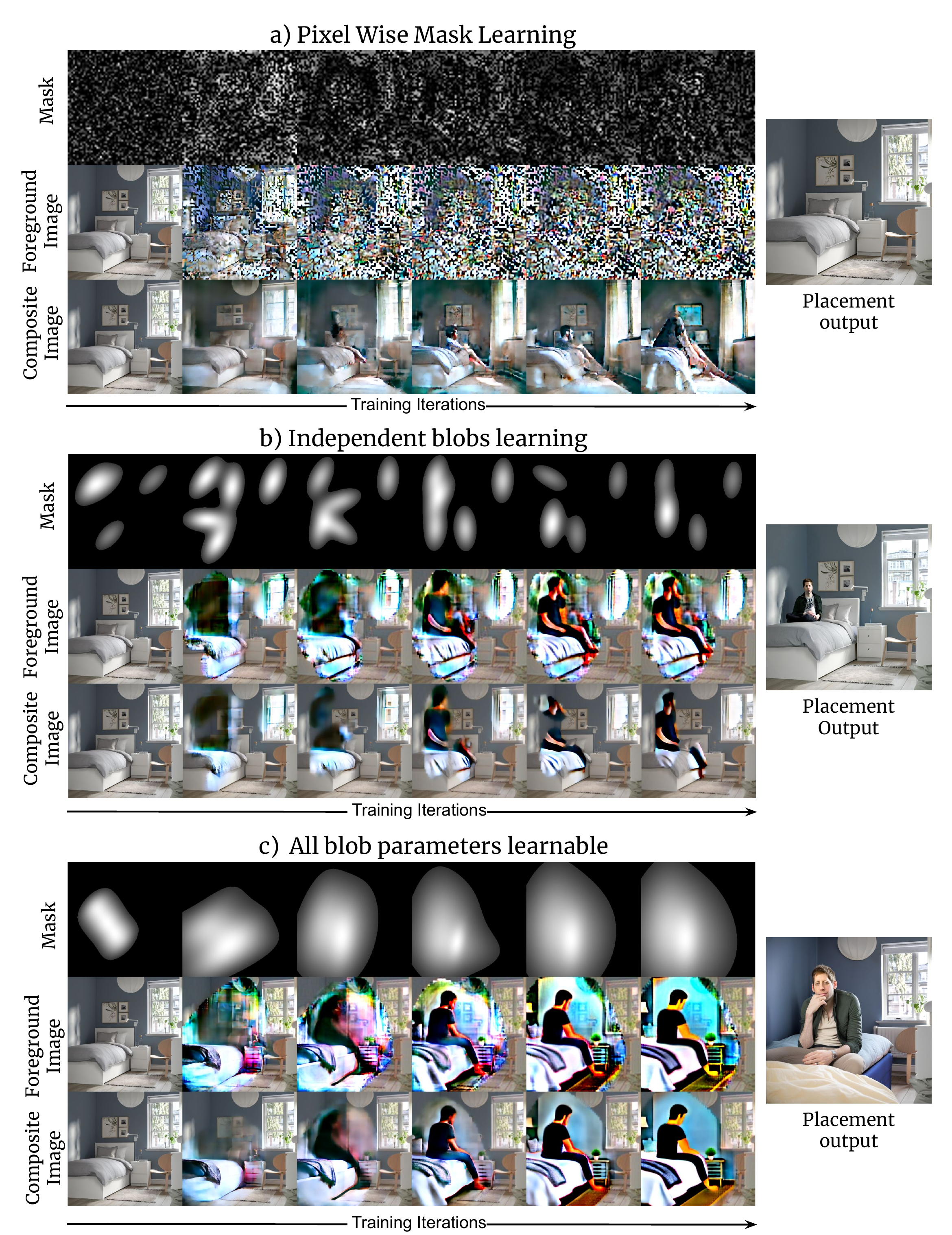}
    \caption{Ablation over different parameterization for \textit{semantic masks}.}
    \vspace{-6mm} 
    \label{fig:blob-baselines}
\end{figure}

\noindent \textbf{b)} Next, we tried parameterizing a mask as a set of Gaussian blobs without tying their centers. During training, the locations of the blobs are updated independently of each other. After convergence, they do not occupy nearby regions to create a continuous \textit{semantic mask} that captures an appropriate human pose. This mask generates the person but changes the background in an unnatural pose. This demonstrates the significance of our design decision to maintain the linkage between the blob centers so that the blobs as a whole can form appropriate human poses and generate plausible semantic masks to place humans. \textbf{c)} We also tried another baseline approach, where all the blob parameters $\mathbf{s}$, $\mathbf{a}$, x $\alpha$, and $\theta$ are learnable. This results in generating masks that have a large foreground region, including large background regions. This is primarily because the blob's scale $\mathbf{s}$ and aspect ratio $\mathbf{a}$ can increase without any regularization to minimize the SDS loss on the combined image. Hence, when used as a semantic mask for the next stage of inpainting, a person with a relatively large scale is generated, which is inconsistent with the background scene or suffers from significant background distortions. This suggests the importance of keeping some of the blob parameters, like aspect ratio and scale, fixed during optimization to obtain plausible masks for realistic placement. 

\clearpage 

\section{Ablations}
\subsection{Semantic mask generation}
\label{sec:supp-ablations-mask}
In this section, we ablate over different design choices for parameterization of \textit{semantic mask}. 

\begin{figure}
    \centering
    \includegraphics[width=0.8\linewidth]{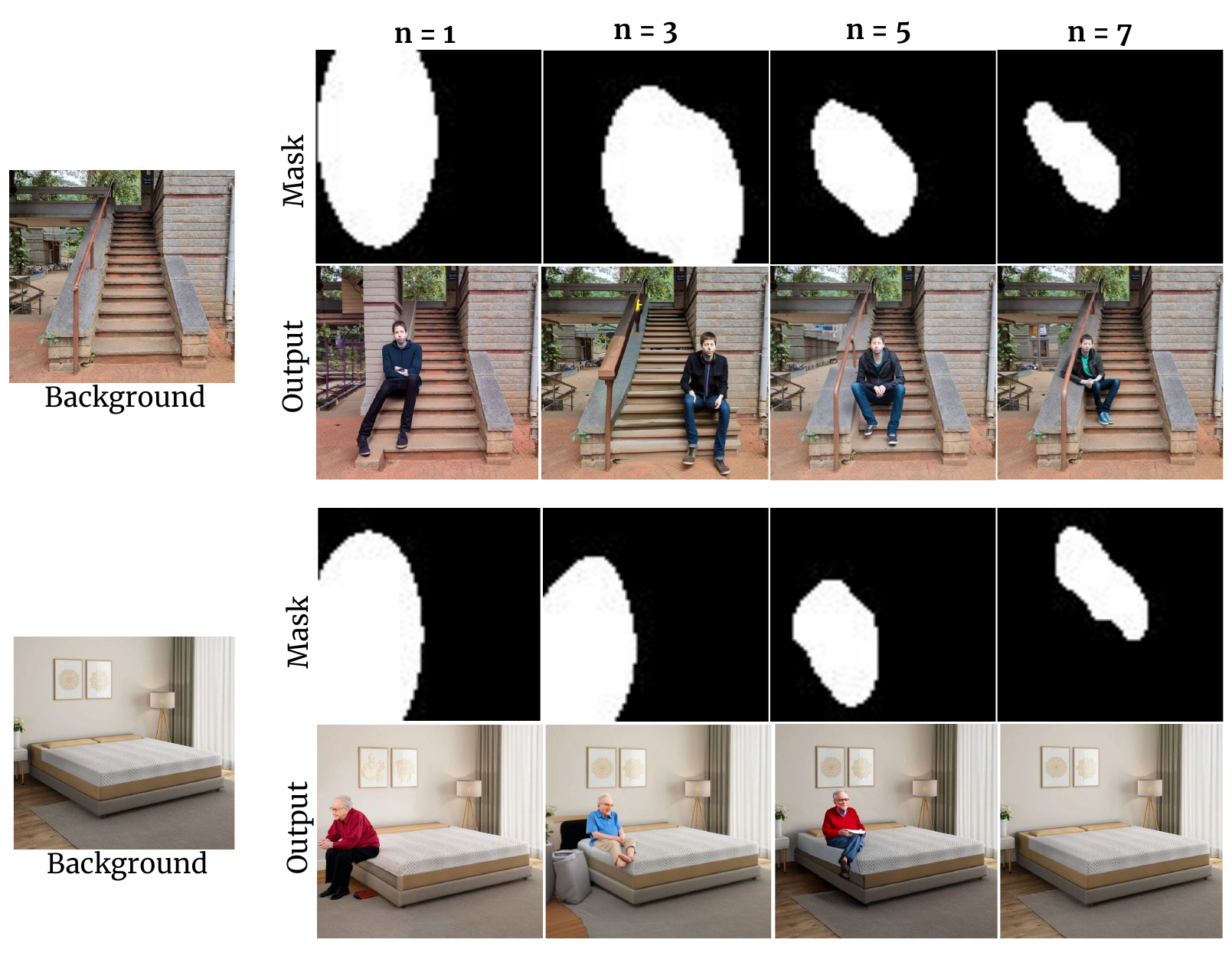}
    \caption{Ablation over number of blobs used to parameterize \textit{semantic mask}.}
    \label{fig:ablate-num-blobs}
\end{figure}

\noindent \textbf{Number of blobs.} We ablate over the number of Gaussian blobs used to parameterize the \textit{semantic mask} in Table.\textcolor{red}{3}b) (main paper). Here, we present qualitative results for the same experiment in Fig.~\ref{fig:ablate-num-blobs}. Using less number of blobs results in a crude semantic mask and doesn't capture the required human pose for accurate placement. This results in significant background changes and changes in the person's pose. Having a very high number with a relatively smaller scale and trying to overfit too much to the person's pose. However, we have observed that the inpainting pipeline works best when we have somewhat loose \textit{semantic masks}, we found $n=5$ achieves the perfect trade-off. We decide the scale based on the number of blobs to give enough space to the mask for generating humans. Specifically, we keep scale values $\{3.0,1.0,0.6,0.43\}$ for number of blobs $\{1,3,5,7\}$ respectively.

\begin{figure}
    \centering
    \includegraphics[width=0.8\linewidth]{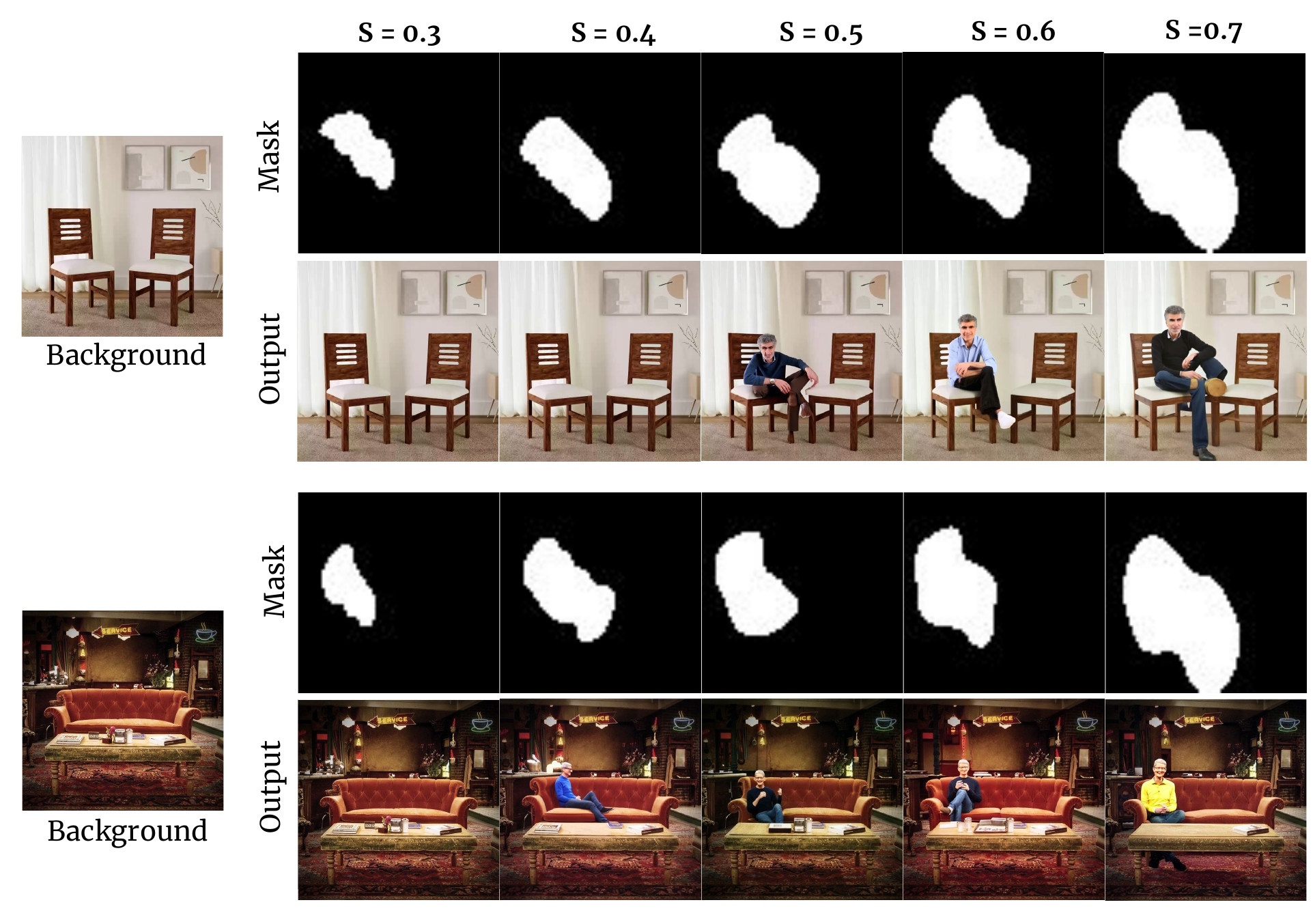}
    \caption{Ablation over the scale of blob $\mathbf{s}$ in \textit{semantic masks}.}
    \label{fig:ablate-blob-scale}
\end{figure}

\clearpage 
\noindent \textbf{Scale of the blobs $\mathbf{s}$.} We present qualitative results for the ablation on blob size from Table.\textcolor{red}{3}a) (main paper) in Fig.~\ref{fig:ablate-blob-scale}. Having a larger scale leads to accurate placement of the person but significantly distorts the background; however, having a small fixed scale for the blob limits the person's size to be inpainted. This results in failure for inpainting where no person is being placed. 

\clearpage

\begin{figure} 
    \centering
    \includegraphics[width=0.8\linewidth]{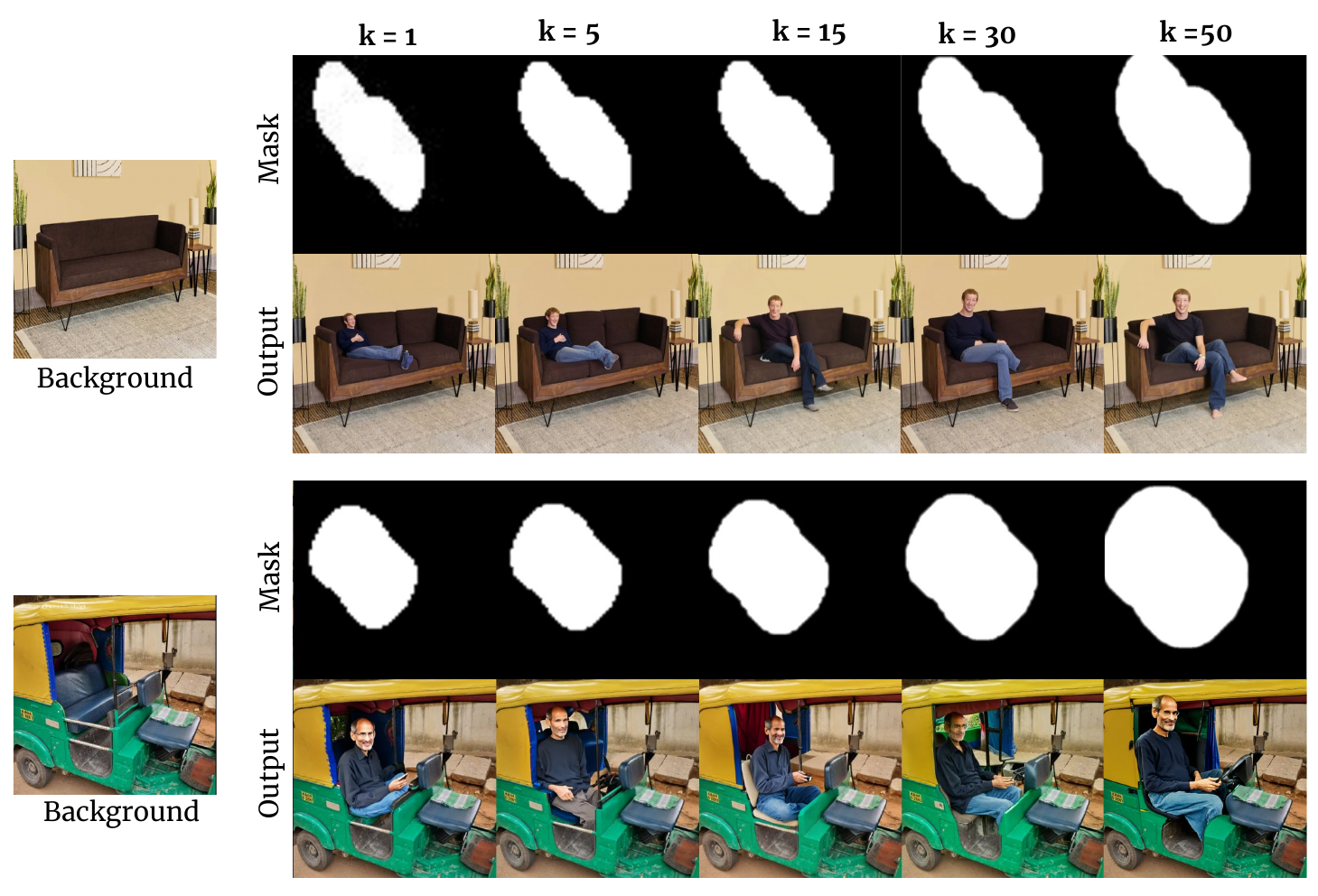}
    \caption{Ablation over the kernel size used dilation of \textit{semantic mask}.}
    \label{fig:ablate-dilate-blob}
\end{figure}

\noindent \textbf{Dilation size.} In our early experiments, we found that using an exact mask of a person for inpainting doesn’t generate the person as shown in Fig.\textcolor{red}{5} (main paper). To this end, we dilated the obtained \textit{semantic mask} after binarization to provide flexibility to the inpainting pipeline. The dilation effect is similar to the blob's scale parameter $\mathbf{s}$. We ablate over the kernel size used for dilation in Fig.~\ref{fig:ablate-dilate-blob}. As we increase the kernel size, the $\%Person$ increases, suggesting successful human placement; however, it increases background distortion. This is also evident in the qualitative results in Fig.~\ref{fig:ablate-dilate-blob}. Having a significantly large kernel size also generates large humans that look unnatural.  

\begin{wraptable}{r}{0.50\textwidth}
\vspace{-10mm}
\caption{Ablation over dilation kernel size.}
\adjustbox{max width=0.5\textwidth}{
\label{tab:ablate-dilate} 
\begin{tabular}{c|c|c|c}
\hline
\textbf{Kernel Size (\textbf{k})} & \textbf{LPIPS $\downarrow$} & \textbf{CLIP-sim $\uparrow$} & \textbf{\% Person $\uparrow$} 
\\ \hline
1                                   & 0.0869         & 0.2699              & 77.9                       \\
3                                   & 0.0882         & 0.2718              & 81.5                       \\
5                                   & 0.0904         & 0.2736              & 90.6                       \\
15                                  & 0.0992         & 0.2735              & 93.1                       \\ 
30                                  & 0.1135         & 0.2749              & 96.8   \\   \hline       

\end{tabular}
}
\vspace{-8mm}
\end{wraptable}
\noindent Due to this, we always dilate our mask obtained in semantic mask generation before we use it for the inpainting task. However, too much dilation could cause a lot of background changes. So we performed ablation studies on the dilation kernel size, and from Tab.~\ref{tab:ablate-dilate}, we find that a kernel size of 15 seems to be performing best with low background change. 

\clearpage 

\subsection{Subject conditioned inpainting}
\label{sec:supp-sub-inpaint}
\begin{figure}[h]
    \centering
    \vspace{-6mm}
    \includegraphics[width=0.95\linewidth]{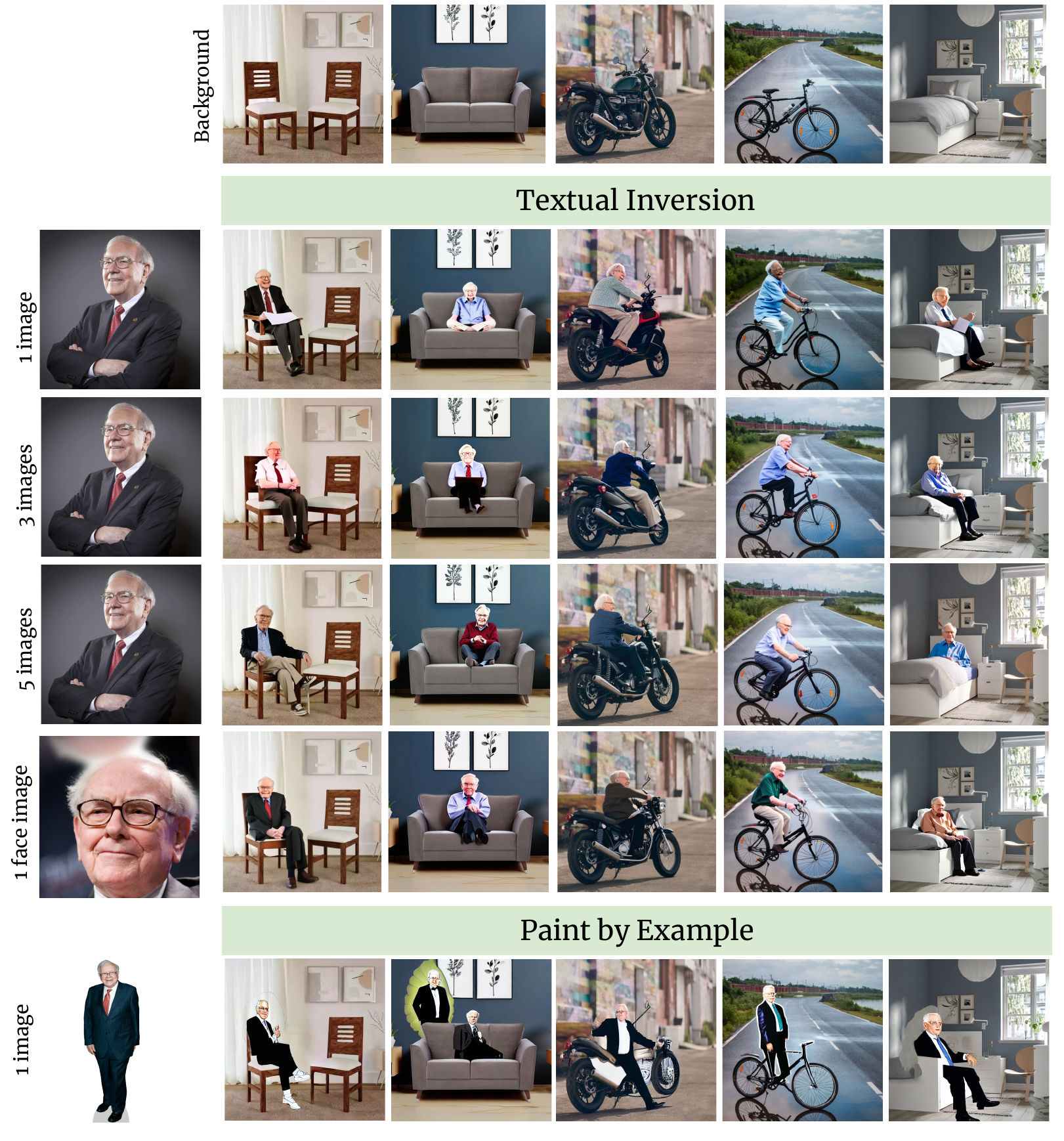}
    \vspace{-4mm}
    \caption{Ablation over various inpainting configurations.}
    \vspace{-6mm}   
    \label{fig:ablate-inpaint}
\end{figure}

We ablate over different inpainting methods to generate plausible human placement results in Fig.~\ref{fig:ablate-inpaint}. For our inpainting pipeline based on Textual Inversion~\cite{textual-inversion}, we use different numbers of training images. Additionally, we used only single-face crop images instead of full-body images to analyze the impact of providing full-body images. Using five images significantly improved the inpainting results with consistent identity in the generation. Using a face image generates decent results, but it has varied body shapes over the generations. We also compared with reference-based inpainting pipeline Paint-by-Example~\cite{paint-by-example}, which generates unnatural outputs with inconsistent identities. Our simplistic method for conditional inpainting generates significantly better human placement results that naturally blend wells in the scene. 

\section{Additional Results.}
\label{sec:addn-details} 

\begin{figure} 
    \centering
    \includegraphics[width=0.95\linewidth]{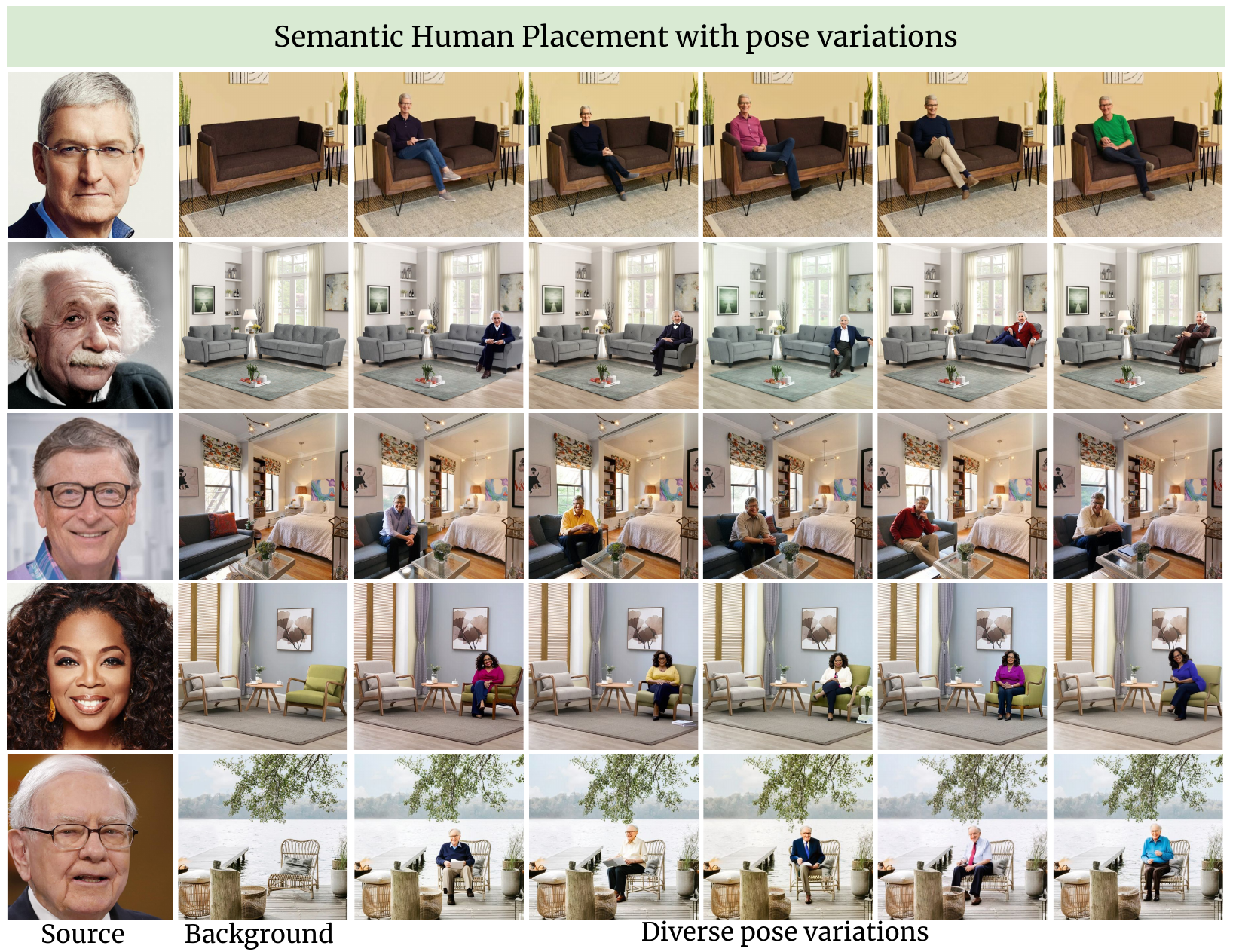}
    \caption{Semantic Human Placement with diverse human poses}
    \label{fig:addn-res1}
\end{figure}

\begin{figure} 
    \centering
    \includegraphics[width=0.95\linewidth]{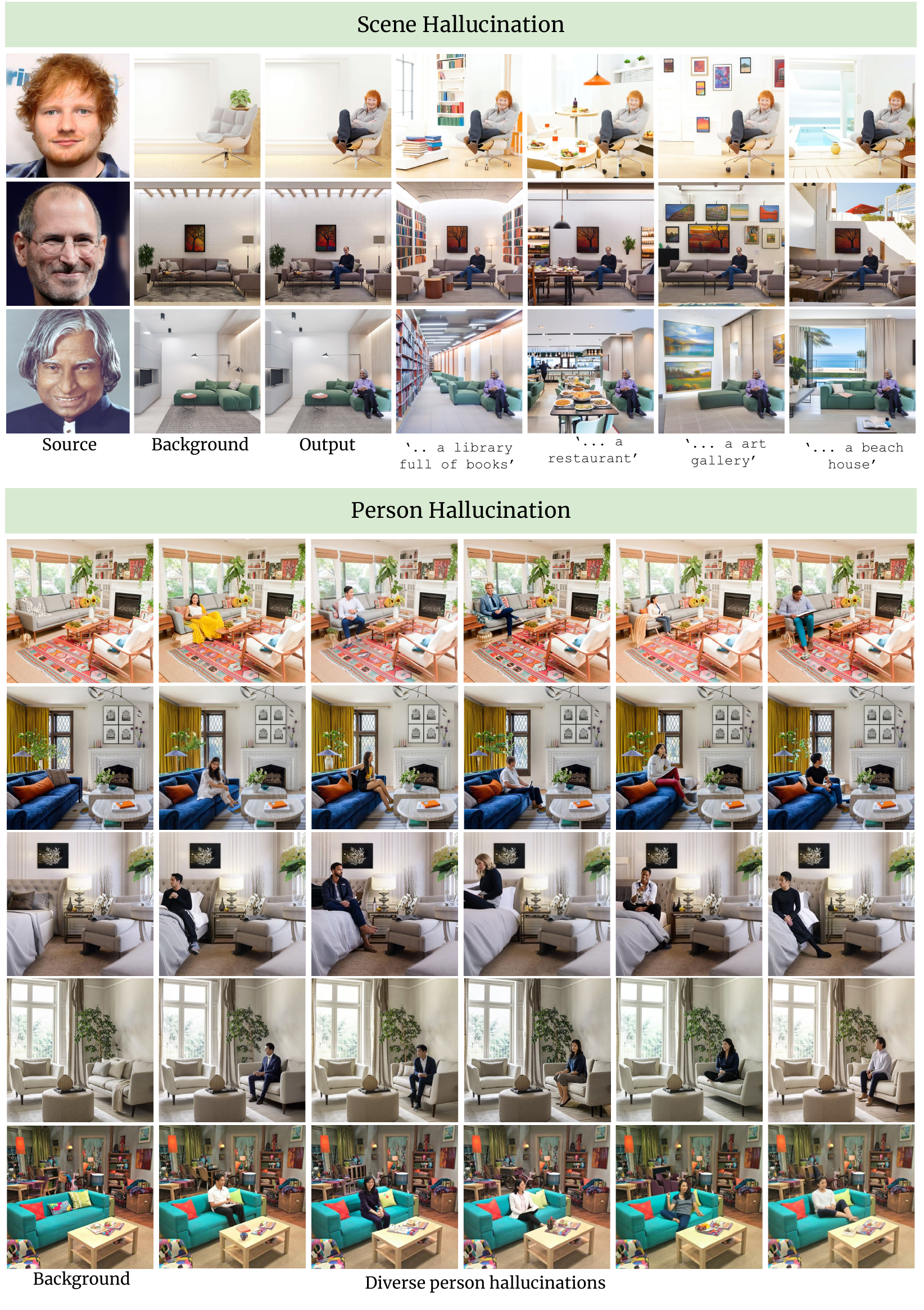}
    \caption{\textbf{Downstream applications} - scene and person hallucinations}
    \label{fig:addn-res2}
\end{figure}

\begin{figure} 
    \centering
    \includegraphics[width=0.95\linewidth]{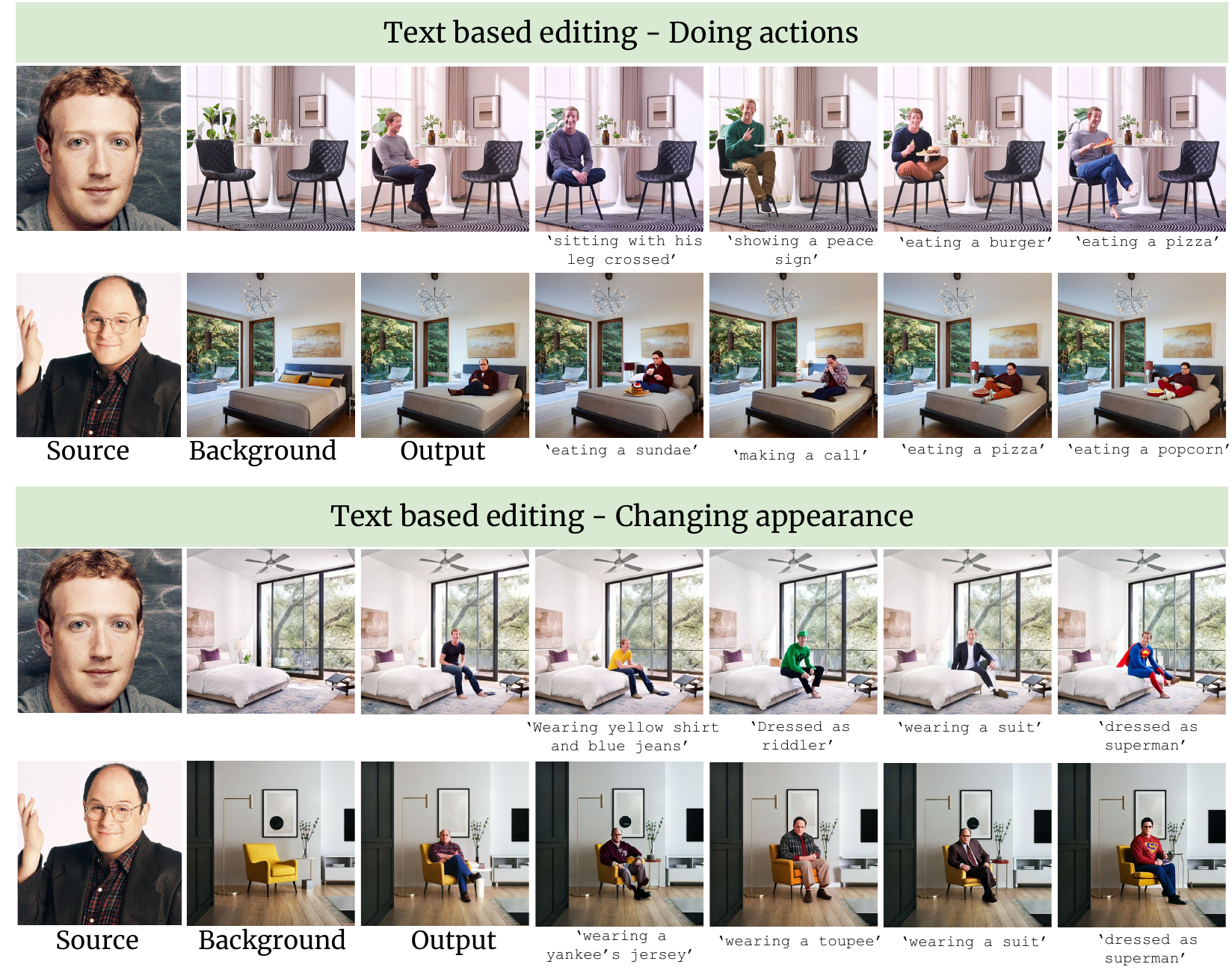}
    \caption{\textbf{Downstream applications} - text-based editing of the generated person in the scene}
    \label{fig:addn-res3}
\end{figure}

We present additional results for Semantic Human Placement, scene and human hallucination and text-based editing from our proposed method in Fig.~\ref{fig:addn-res1},~\ref{fig:addn-res2},~\ref{fig:addn-res3}.

\clearpage 

\section{Prompts used for Placing Person}
\begin{itemize}
\item A person sitting on a bed
\item A person sitting on a sofa
\item A person sitting on a chair
\item A person sitting on a bench
\item A person sitting in a car
\item A person sitting on stairs
\item A person sitting in an auto rickshaw
\item A person riding a motorbike
\item A person riding a cycle
\item A person playing pool
\item A person walking on a road
\item A person sitting there
\item A person running there
\item A person walking there
\item A person riding a cycle
\item A person walking on a sidewalk
\item A person standing on a podium
\item A person standing in a room
\item A person standing outdoor
\item A person standing near the Eiffel Tower
\item A person standing in front of the Taj Mahal
\end{itemize}

%
%

%
%
\bibliographystyle{splncs04}
\bibliography{main}

\begin{thebibliography}{10}
\providecommand{\url}[1]{\texttt{#1}}
\providecommand{\urlprefix}{URL }
\providecommand{\doi}[1]{https://doi.org/#1}

\bibitem{achiam2023gpt}
Achiam, J., Adler, S., Agarwal, S., Ahmad, L., Akkaya, I., Aleman, F.L., Almeida, D., Altenschmidt, J., Altman, S., Anadkat, S., et~al.: Gpt-4 technical report. arXiv preprint arXiv:2303.08774  (2023)

\bibitem{old_inp1}
Ballester, C., Bertalmio, M., Caselles, V., Sapiro, G., Verdera, J.: Filling-in by joint interpolation of vector fields and gray levels. IEEE transactions on image processing  \textbf{10}(8),  1200--1211 (2001)

\bibitem{trans_inp1}
Bar, A., Gandelsman, Y., Darrell, T., Globerson, A., Efros, A.: Visual prompting via image inpainting. Advances in Neural Information Processing Systems  \textbf{35},  25005--25017 (2022)

\bibitem{old_inp2}
Barnes, C., Shechtman, E., Finkelstein, A., Goldman, D.B.: Patchmatch: A randomized correspondence algorithm for structural image editing. ACM Trans. Graph.  \textbf{28}(3), ~24 (2009)

\bibitem{brooks2022hallucinating}
Brooks, T., Efros, A.A.: Hallucinating pose-compatible scenes. In: European Conference on Computer Vision. pp. 510--528. Springer (2022)

\bibitem{cao2020long}
Cao, Z., Gao, H., Mangalam, K., Cai, Q.Z., Vo, M., Malik, J.: Long-term human motion prediction with scene context. In: Computer Vision--ECCV 2020: 16th European Conference, Glasgow, UK, August 23--28, 2020, Proceedings, Part I 16. pp. 387--404. Springer (2020)

\bibitem{cao2021reconstructing}
Cao, Z., Radosavovic, I., Kanazawa, A., Malik, J.: Reconstructing hand-object interactions in the wild. In: Proceedings of the IEEE/CVF International Conference on Computer Vision. pp. 12417--12426 (2021)

\bibitem{attend-excite}
Chefer, H., Alaluf, Y., Vinker, Y., Wolf, L., Cohen-Or, D.: Attend-and-excite: Attention-based semantic guidance for text-to-image diffusion models. ACM Transactions on Graphics (TOG)  \textbf{42}(4),  1--10 (2023)

\bibitem{chen2023anydoor}
Chen, X., Huang, L., Liu, Y., Shen, Y., Zhao, D., Zhao, H.: Anydoor: Zero-shot object-level image customization. arXiv preprint arXiv:2307.09481  (2023)

\bibitem{Chen2023Textto3DUG}
Chen, Z., Wang, F., Liu, H.: Text-to-3d using gaussian splatting. ArXiv  (2023)

\bibitem{chuang2018learning}
Chuang, C.Y., Li, J., Torralba, A., Fidler, S.: Learning to act properly: Predicting and explaining affordances from images. In: Proceedings of the IEEE Conference on Computer Vision and Pattern Recognition. pp. 975--983 (2018)

\bibitem{direct-affordance3}
Delaitre, V., Fouhey, D.F., Laptev, I., Sivic, J., Gupta, A., Efros, A.A.: Scene semantics from long-term observation of people. In: Computer Vision--ECCV 2012: 12th European Conference on Computer Vision, Florence, Italy, October 7-13, 2012, Proceedings, Part VI 12. pp. 284--298. Springer (2012)

\bibitem{delaitre2012scene}
Delaitre, V., Fouhey, D.F., Laptev, I., Sivic, J., Gupta, A., Efros, A.A.: Scene semantics from long-term observation of people. In: Computer Vision--ECCV 2012: 12th European Conference on Computer Vision, Florence, Italy, October 7-13, 2012, Proceedings, Part VI 12. pp. 284--298. Springer (2012)

\bibitem{diffusion-beats-gan}
Dhariwal, P., Nichol, A.: Diffusion models beat gans on image synthesis. Advances in Neural Information Processing Systems  \textbf{34},  8780--8794 (2021)

\bibitem{old_inp4}
Efros, A.A., Leung, T.K.: Texture synthesis by non-parametric sampling. In: Proceedings of the seventh IEEE international conference on computer vision. vol.~2, pp. 1033--1038. IEEE (1999)

\bibitem{epstein2022blobgan}
Epstein, D., Park, T., Zhang, R., Shechtman, E., Efros, A.A.: Blobgan: Spatially disentangled scene representations. In: European Conference on Computer Vision. pp. 616--635. Springer (2022)

\bibitem{trans_inp2}
Esser, P., Rombach, R., Ommer, B.: Taming transformers for high-resolution image synthesis. In: Proceedings of the IEEE/CVF conference on computer vision and pattern recognition. pp. 12873--12883 (2021)

\bibitem{direct-affordance1}
Fouhey, D.F., Wang, X., Gupta, A.: In defense of the direct perception of affordances. arXiv preprint arXiv:1505.01085  (2015)

\bibitem{fouhey2015defense}
Fouhey, D.F., Wang, X., Gupta, A.: In defense of the direct perception of affordances. arXiv preprint arXiv:1505.01085  (2015)

\bibitem{textual-inversion}
Gal, R., Alaluf, Y., Atzmon, Y., Patashnik, O., Bermano, A.H., Chechik, G., Cohen-Or, D.: An image is worth one word: Personalizing text-to-image generation using textual inversion. arXiv preprint arXiv:2208.01618  (2022)

\bibitem{e4t}
Gal, R., Arar, M., Atzmon, Y., Bermano, A.H., Chechik, G., Cohen-Or, D.: Encoder-based domain tuning for fast personalization of text-to-image models. ACM Transactions on Graphics (TOG)  \textbf{42}(4),  1--13 (2023)

\bibitem{gibson1978ecological}
Gibson, J.J.: The ecological approach to the visual perception of pictures. Leonardo  \textbf{11}(3),  227--235 (1978)

\bibitem{gkioxari2018detecting}
Gkioxari, G., Girshick, R., Doll{\'a}r, P., He, K.: Detecting and recognizing human-object interactions. In: Proceedings of the IEEE conference on computer vision and pattern recognition. pp. 8359--8367 (2018)

\bibitem{grabner2011makes}
Grabner, H., Gall, J., Van~Gool, L.: What makes a chair a chair? In: CVPR 2011. pp. 1529--1536. IEEE (2011)

\bibitem{gupta20113d}
Gupta, A., Satkin, S., Efros, A.A., Hebert, M.: From 3d scene geometry to human workspace. In: CVPR 2011. pp. 1961--1968. IEEE (2011)

\bibitem{dds}
Hertz, A., Aberman, K., Cohen-Or, D.: Delta denoising score. In: Proceedings of the IEEE/CVF International Conference on Computer Vision. pp. 2328--2337 (2023)

\bibitem{prompt2prompt}
Hertz, A., Mokady, R., Tenenbaum, J., Aberman, K., Pritch, Y., Cohen-or, D.: Prompt-to-prompt image editing with cross-attention control. In: The Eleventh International Conference on Learning Representations (2022)

\bibitem{ddpm}
Ho, J., Jain, A., Abbeel, P.: Denoising diffusion probabilistic models. Advances in Neural Information Processing Systems  \textbf{33},  6840--6851 (2020)

\bibitem{huggingface2022diffusers}
HuggingFace: Diffusers. \url{https://github.com/huggingface/diffusers} (2022)

\bibitem{isola2017image}
Isola, P., Zhu, J.Y., Zhou, T., Efros, A.A.: Image-to-image translation with conditional adversarial networks. In: Proceedings of the IEEE conference on computer vision and pattern recognition. pp. 1125--1134 (2017)

\bibitem{jiang2013hallucinated}
Jiang, Y., Koppula, H., Saxena, A.: Hallucinated humans as the hidden context for labeling 3d scenes. In: Proceedings of the IEEE Conference on Computer Vision and Pattern Recognition. pp. 2993--3000 (2013)

\bibitem{noisefreeSDS}
Katzir, O., Patashnik, O., Cohen-Or, D., Lischinski, D.: Noise-free score distillation. arXiv preprint arXiv:2310.17590  (2023)

\bibitem{imagic}
Kawar, B., Zada, S., Lang, O., Tov, O., Chang, H., Dekel, T., Mosseri, I., Irani, M.: Imagic: Text-based real image editing with diffusion models. In: Proceedings of the IEEE/CVF Conference on Computer Vision and Pattern Recognition. pp. 6007--6017 (2023)

\bibitem{sam}
Kirillov, A., Mintun, E., Ravi, N., Mao, H., Rolland, C., Gustafson, L., Xiao, T., Whitehead, S., Berg, A.C., Lo, W.Y., et~al.: Segment anything. arXiv preprint arXiv:2304.02643  (2023)

\bibitem{koppula2013learning}
Koppula, H.S., Gupta, R., Saxena, A.: Learning human activities and object affordances from rgb-d videos. The International journal of robotics research  \textbf{32}(8),  951--970 (2013)

\bibitem{kulal2023putting}
Kulal, S., Brooks, T., Aiken, A., Wu, J., Yang, J., Lu, J., Efros, A.A., Singh, K.K.: Putting people in their place: Affordance-aware human insertion into scenes. In: Proceedings of the IEEE/CVF Conference on Computer Vision and Pattern Recognition. pp. 17089--17099 (2023)

\bibitem{custom-diffusion}
Kumari, N., Zhang, B., Zhang, R., Shechtman, E., Zhu, J.Y.: Multi-concept customization of text-to-image diffusion. In: Proceedings of the IEEE/CVF Conference on Computer Vision and Pattern Recognition. pp. 1931--1941 (2023)

\bibitem{lee2002interactive}
Lee, J., Chai, J., Reitsma, P.S., Hodgins, J.K., Pollard, N.S.: Interactive control of avatars animated with human motion data. In: Proceedings of the 29th annual conference on Computer graphics and interactive techniques. pp. 491--500 (2002)

\bibitem{li2019putting}
Li, X., Liu, S., Kim, K., Wang, X., Yang, M.H., Kautz, J.: Putting humans in a scene: Learning affordance in 3d indoor environments. In: Proceedings of the IEEE/CVF Conference on Computer Vision and Pattern Recognition. pp. 12368--12376 (2019)

\bibitem{llava}
Liu, H., Li, C., Li, Y., Lee, Y.J.: Improved baselines with visual instruction tuning. arXiv preprint arXiv:2310.03744  (2023)

\bibitem{opa-dataset}
Liu, L., Liu, Z., Zhang, B., Li, J., Niu, L., Liu, Q., Zhang, L.: Opa: object placement assessment dataset. arXiv preprint arXiv:2107.01889  (2021)

\bibitem{multi_image_guide_inp}
Lu, L., Zhang, B., Niu, L.: Dreamcom: Finetuning text-guided inpainting model for image composition. arXiv preprint arXiv:2309.15508  (2023)

\bibitem{diff_inp1}
Lugmayr, A., Danelljan, M., Romero, A., Yu, F., Timofte, R., Van~Gool, L.: Repaint: Inpainting using denoising diffusion probabilistic models. In: Proceedings of the IEEE/CVF Conference on Computer Vision and Pattern Recognition. pp. 11461--11471 (2022)

\bibitem{meng2021sdedit}
Meng, C., He, Y., Song, Y., Song, J., Wu, J., Zhu, J.Y., Ermon, S.: Sdedit: Guided image synthesis and editing with stochastic differential equations. arXiv preprint arXiv:2108.01073  (2021)

\bibitem{old_inp3}
Osher, S., Burger, M., Goldfarb, D., Xu, J., Yin, W.: An iterative regularization method for total variation-based image restoration. Multiscale Modeling \& Simulation  \textbf{4}(2),  460--489 (2005)

\bibitem{parihar2022everything}
Parihar, R., Dhiman, A., Karmali, T., Babu, R.V.: Everything is there in latent space: Attribute editing and attribute style manipulation by stylegan latent space exploration. In: Proceedings of the 30th ACM International Conference on Multimedia. pp. 1828--1836 (2022)

\bibitem{motionstyle}
Parihar, R., Magazine, R., Tiwari, P., Babu, R.V.: We never go out of style: Motion disentanglement by subspace decomposition of latent space. arXiv preprint arXiv:2306.00559  (2023)

\bibitem{park2020swapping}
Park, T., Zhu, J.Y., Wang, O., Lu, J., Shechtman, E., Efros, A., Zhang, R.: Swapping autoencoder for deep image manipulation. Advances in Neural Information Processing Systems  \textbf{33},  7198--7211 (2020)

\bibitem{zeroshotimg2img}
Parmar, G., Kumar~Singh, K., Zhang, R., Li, Y., Lu, J., Zhu, J.Y.: Zero-shot image-to-image translation. In: ACM SIGGRAPH 2023 Conference Proceedings. pp. 1--11 (2023)

\bibitem{sds}
Poole, B., Jain, A., Barron, J.T., Mildenhall, B.: Dreamfusion: Text-to-3d using 2d diffusion. arXiv preprint arXiv:2209.14988  (2022)

\bibitem{quadflieg2017neuroscience}
Quadflieg, S., Koldewyn, K.: The neuroscience of people watching: how the human brain makes sense of other people's encounters. Annals of the New York Academy of Sciences  \textbf{1396}(1),  166--182 (2017)

\bibitem{clip}
Radford, A., Kim, J.W., Hallacy, C., Ramesh, A., Goh, G., Agarwal, S., Sastry, G., Askell, A., Mishkin, P., Clark, J., et~al.: Learning transferable visual models from natural language supervision. In: International conference on machine learning. pp. 8748--8763. PMLR (2021)

\bibitem{dalle-2}
Ramesh, A., Dhariwal, P., Nichol, A., Chu, C., Chen, M.: Hierarchical text-conditional image generation with clip latents. arXiv preprint arXiv:2204.06125  \textbf{1}(2), ~3 (2022)

\bibitem{dalle}
Ramesh, A., Pavlov, M., Goh, G., Gray, S., Voss, C., Radford, A., Chen, M., Sutskever, I.: Zero-shot text-to-image generation. In: International Conference on Machine Learning. pp. 8821--8831. PMLR (2021)

\bibitem{seeing-the-unseen}
Ramrakhya, R., Kembhavi, A., Batra, D., Kira, Z., Zeng, K.H., Weihs, L.: Seeing the unseen: Visual common sense for semantic placement. arXiv preprint arXiv:2401.07770  (2024)

\bibitem{roettger2021xltextualinversion}
Roettger, T.: Xl-textual-inversion. \url{https://github.com/oss-roettger/XL-Textual-Inversion/blob/main/XL_Inversion.ipynb} (2023)

\bibitem{ldm}
Rombach, R., Blattmann, A., Lorenz, D., Esser, P., Ommer, B.: High-resolution image synthesis with latent diffusion models. In: Proceedings of the IEEE/CVF Conference on Computer Vision and Pattern Recognition. pp. 10684--10695 (2022)

\bibitem{ruiz2023dreambooth}
Ruiz, N., Li, Y., Jampani, V., Pritch, Y., Rubinstein, M., Aberman, K.: Dreambooth: Fine tuning text-to-image diffusion models for subject-driven generation. In: Proceedings of the IEEE/CVF Conference on Computer Vision and Pattern Recognition. pp. 22500--22510 (2023)

\bibitem{diff_inp2}
Saharia, C., Chan, W., Chang, H., Lee, C., Ho, J., Salimans, T., Fleet, D., Norouzi, M.: Palette: Image-to-image diffusion models. In: ACM SIGGRAPH 2022 Conference Proceedings. pp. 1--10 (2022)

\bibitem{imagen}
Saharia, C., Chan, W., Saxena, S., Li, L., Whang, J., Denton, E.L., Ghasemipour, K., Gontijo~Lopes, R., Karagol~Ayan, B., Salimans, T., et~al.: Photorealistic text-to-image diffusion models with deep language understanding. Advances in Neural Information Processing Systems  \textbf{35},  36479--36494 (2022)

\bibitem{shen2020interpreting}
Shen, Y., Gu, J., Tang, X., Zhou, B.: Interpreting the latent space of gans for semantic face editing. In: Proceedings of the IEEE/CVF conference on computer vision and pattern recognition. pp. 9243--9252 (2020)

\bibitem{singh2024smartmask}
Singh, J., Zhang, J., Liu, Q., Smith, C., Lin, Z., Zheng, L.: Smartmask: Context aware high-fidelity mask generation for fine-grained object insertion and layout control. In: Proceedings of the IEEE/CVF Conference on Computer Vision and Pattern Recognition. pp. 6497--6506 (2024)

\bibitem{ddim}
Song, J., Meng, C., Ermon, S.: Denoising diffusion implicit models. arXiv preprint arXiv:2010.02502  (2020)

\bibitem{tang2023dreamgaussian}
Tang, J., Ren, J., Zhou, H., Liu, Z., Zeng, G.: Dreamgaussian: Generative gaussian splatting for efficient 3d content creation. arXiv preprint arXiv:2309.16653  (2023)

\bibitem{tumanyan2022plugandplay}
Tumanyan, N., Geyer, M., Bagon, S., Dekel, T.: Plug-and-play diffusion features for text-driven image-to-image translation. In: Proceedings of the IEEE/CVF Conference on Computer Vision and Pattern Recognition. pp. 1921--1930 (2023)

\bibitem{wang2021synthesizing}
Wang, J., Xu, H., Xu, J., Liu, S., Wang, X.: Synthesizing long-term 3d human motion and interaction in 3d scenes. In: Proceedings of the IEEE/CVF Conference on Computer Vision and Pattern Recognition. pp. 9401--9411 (2021)

\bibitem{Wang2024InstantIDZI}
Wang, Q., Bai, X., Wang, H., Qin, Z., Chen, A.: Instantid: Zero-shot identity-preserving generation in seconds. ArXiv  (2024)

\bibitem{direct-affordanc2-binge}
Wang, X., Girdhar, R., Gupta, A.: Binge watching: Scaling affordance learning from sitcoms. In: Proceedings of the IEEE Conference on Computer Vision and Pattern Recognition. pp. 2596--2605 (2017)

\bibitem{wang2017binge}
Wang, X., Girdhar, R., Gupta, A.: Binge watching: Scaling affordance learning from sitcoms. In: Proceedings of the IEEE Conference on Computer Vision and Pattern Recognition. pp. 2596--2605 (2017)

\bibitem{wang2021nerfmm}
Wang, Z., Wu, S., Xie, W., Chen, M., Prisacariu, V.A.: Ne{RF}$--$: Neural radiance fields without known camera parameters. arXiv preprint arXiv:2102.07064  (2021)

\bibitem{diffusion-im2im}
Wu, C.H., De~la Torre, F.: Unifying diffusion models' latent space, with applications to cyclediffusion and guidance. arXiv preprint arXiv:2210.05559  (2022)

\bibitem{single_image_guide_inp}
Xie, S., Zhao, Y., Xiao, Z., Chan, K.C., Li, Y., Xu, Y., Zhang, K., Hou, T.: Dreaminpainter: Text-guided subject-driven image inpainting with diffusion models. arXiv preprint arXiv:2312.03771  (2023)

\bibitem{paint-by-example}
Yang, B., Gu, S., Zhang, B., Zhang, T., Chen, X., Sun, X., Chen, D., Wen, F.: Paint by example: Exemplar-based image editing with diffusion models. In: Proceedings of the IEEE/CVF Conference on Computer Vision and Pattern Recognition. pp. 18381--18391 (2023)

\bibitem{cnn_inp1}
Yang, C., Lu, X., Lin, Z., Shechtman, E., Wang, O., Li, H.: High-resolution image inpainting using multi-scale neural patch synthesis. In: Proceedings of the IEEE conference on computer vision and pattern recognition. pp. 6721--6729 (2017)

\bibitem{single_image_guide}
Yang, S., Chen, X., Liao, J.: Uni-paint: A unified framework for multimodal image inpainting with pretrained diffusion model. In: Proceedings of the 31st ACM International Conference on Multimedia. pp. 3190--3199 (2023)

\bibitem{yao2010modeling}
Yao, B., Fei-Fei, L.: Modeling mutual context of object and human pose in human-object interaction activities. In: 2010 IEEE Computer Society Conference on Computer Vision and Pattern Recognition. pp. 17--24. IEEE (2010)

\bibitem{ip-adap}
Ye, H., Zhang, J., Liu, S., Han, X., Yang, W.: Ip-adapter: Text compatible image prompt adapter for text-to-image diffusion models. arXiv preprint arXiv:2308.06721  (2023)

\bibitem{Ye2023IPAdapterTC}
Ye, H., Zhang, J., Liu, S., Han, X., Yang, W.: Ip-adapter: Text compatible image prompt adapter for text-to-image diffusion models. ArXiv  \textbf{abs/2308.06721} (2023)

\bibitem{cnn_inp2}
Yu, J., Lin, Z., Yang, J., Shen, X., Lu, X., Huang, T.S.: Generative image inpainting with contextual attention. In: Proceedings of the IEEE conference on computer vision and pattern recognition. pp. 5505--5514 (2018)

\bibitem{cnn_inp3}
Yu, J., Lin, Z., Yang, J., Shen, X., Lu, X., Huang, T.S.: Free-form image inpainting with gated convolution. In: Proceedings of the IEEE/CVF international conference on computer vision. pp. 4471--4480 (2019)

\bibitem{trans_inp3}
Yu, Y., Zhan, F., Wu, R., Pan, J., Cui, K., Lu, S., Ma, F., Xie, X., Miao, C.: Diverse image inpainting with bidirectional and autoregressive transformers. In: Proceedings of the 29th ACM International Conference on Multimedia. pp. 69--78 (2021)

\bibitem{celeb-basis}
Yuan, G., Cun, X., Zhang, Y., Li, M., Qi, C., Wang, X., Shan, Y., Zheng, H.: Inserting anybody in diffusion models via celeb basis. arXiv preprint arXiv:2306.00926  (2023)

\bibitem{zhang2023adding_controlnet}
Zhang, L., Agrawala, M.: Adding conditional control to text-to-image diffusion models pp. 3836--3847 (2023)

\bibitem{zhang2018unreasonable}
Zhang, R., Isola, P., Efros, A.A., Shechtman, E., Wang, O.: The unreasonable effectiveness of deep features as a perceptual metric. In: Proceedings of the IEEE conference on computer vision and pattern recognition. pp. 586--595 (2018)

\bibitem{zhang2023paste}
Zhang, X., Guo, J., Yoo, P., Matsuo, Y., Iwasawa, Y.: Paste, inpaint and harmonize via denoising: Subject-driven image editing with pre-trained diffusion model. arXiv preprint arXiv:2306.07596  (2023)

\bibitem{cnn_inp4}
Zhao, S., Cui, J., Sheng, Y., Dong, Y., Liang, X., Chang, E.I., Xu, Y.: Large scale image completion via co-modulated generative adversarial networks. arXiv preprint arXiv:2103.10428  (2021)

\bibitem{cnn_inp5}
Zheng, H., Lin, Z., Lu, J., Cohen, S., Shechtman, E., Barnes, C., Zhang, J., Xu, N., Amirghodsi, S., Luo, J.: Image inpainting with cascaded modulation gan and object-aware training. In: European Conference on Computer Vision. pp. 277--296. Springer (2022)

\bibitem{graconet}
Zhou, S., Liu, L., Niu, L., Zhang, L.: Learning object placement via dual-path graph completion. In: European Conference on Computer Vision. pp. 373--389. Springer (2022)

\bibitem{profusion}
Zhou, Y., Zhang, R., Sun, T., Xu, J.: Enhancing detail preservation for customized text-to-image generation: A regularization-free approach. arXiv preprint arXiv:2305.13579  (2023)

\bibitem{topnet}
Zhu, S., Lin, Z., Cohen, S., Kuen, J., Zhang, Z., Chen, C.: Topnet: Transformer-based object placement network for image compositing. In: Proceedings of the IEEE/CVF Conference on Computer Vision and Pattern Recognition. pp. 1838--1847 (2023)

\bibitem{zhu2023topnet}
Zhu, S., Lin, Z., Cohen, S., Kuen, J., Zhang, Z., Chen, C.: Topnet: Transformer-based object placement network for image compositing. In: Proceedings of the IEEE/CVF Conference on Computer Vision and Pattern Recognition. pp. 1838--1847 (2023)

\bibitem{zhu2014reasoning}
Zhu, Y., Fathi, A., Fei-Fei, L.: Reasoning about object affordances in a knowledge base representation. In: Computer Vision--ECCV 2014: 13th European Conference, Zurich, Switzerland, September 6-12, 2014, Proceedings, Part II 13. pp. 408--424. Springer (2014)

\end{thebibliography}
\end{document}